%% file: iclr2026_conference.tex
\PassOptionsToPackage{breaklinks,colorlinks,allcolors=black}{hyperref}
\documentclass{article} 
\usepackage{iclr2026_conference,times}
\usepackage[pdfcreator={Unknown},pdfproducer={Unknown}]{hyperref}

\input{math_commands.tex}

\usepackage{hyperref}
\usepackage{url}
\usepackage{booktabs}
\usepackage{multirow}
\usepackage[table]{xcolor}
\usepackage{pifont}
\usepackage{graphicx}
\usepackage{pgfplots}
\pgfplotsset{compat=1.18}
\usepackage{cleveref}
\usepackage{subcaption}
\usepackage{tikz}
\usetikzlibrary{positioning, shapes.geometric, backgrounds}
\usepackage{adjustbox}
\usepackage{xcolor}
\usepackage{svg}
\svgpath{{}}

\usepackage{fontawesome5} 

\title{The Deleuzian Representation Hypothesis}

\author{Clément Cornet, Romaric Besançon \& Hervé Le Borgne \\ 
 Université Paris-Saclay, CEA, List,\\
 F-91120, Palaiseau, France  \\
   \texttt{\{clement.cornet,romaric.besancon,herve.le-borgne\}@cea.fr}
}

\renewcommand{\cite}[1]{\citep{#1}}

\colorlet{revcolor}{orange}

\iclrfinalcopy

\begin{document}
\maketitle

\begin{abstract}
We propose an alternative to sparse autoencoders (SAEs) as a simple and effective unsupervised method for extracting interpretable concepts from neural networks. The core idea is to cluster differences in activations, which we formally justify within a discriminant analysis framework. To enhance the diversity of extracted concepts, we refine the approach by weighting the clustering using the skewness of activations. The method aligns with Deleuze's modern view of concepts as differences. We evaluate the approach across five models and three modalities (vision, language, and audio), measuring concept quality, diversity, and consistency. Our results show that the proposed method achieves concept quality surpassing prior unsupervised SAE variants while approaching supervised baselines, and that the extracted concepts enable steering of a model’s inner representations, demonstrating their causal influence on downstream behavior.
\end{abstract}

\begin{figure}[b]
  \centering
  
  \newlength{\boxwidth}
  \setlength{\boxwidth}{0.3\textwidth}
  \newlength{\boxheight}
  \setlength{\boxheight}{4cm}

  \begin{subfigure}[b]{\boxwidth}
    \centering
    \fbox{\parbox[c][\boxheight][c]{\linewidth}{\centering \includegraphics[height=\boxheight]{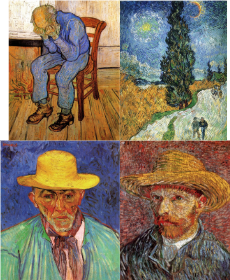}}}
    \caption{Image: Van Gogh's Paintings}
  \end{subfigure}
  \hfill
  \begin{subfigure}[b]{\boxwidth}
    \centering
    \fbox{%
      \begin{minipage}[c][\boxheight][t]{\linewidth}
        \vspace{0.5cm}
        \hspace{0.1\linewidth} ``Winning the prize'' \par
        \vspace{1.0cm}
        \hspace{0.35\linewidth} ``the Gold Medal'' \par
        \vspace{1.0cm}
        \hspace{0.1\linewidth} ``the World Record''
      \end{minipage}
    }
    \caption{Text: Sports Achievements}
  \end{subfigure}
  \hfill
  \begin{subfigure}[b]{\boxwidth}
    \centering
    \fbox{\parbox[c][\boxheight][c]{\linewidth}{
    \centering
    \includegraphics[width=\boxwidth]{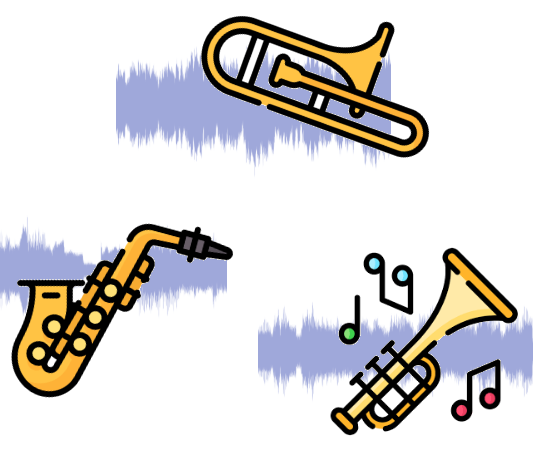}}
    }
    \caption{Audio: Brass Instruments}
  \end{subfigure}
  
  \caption{Our method extracts diverse concepts from image, text and audio models.
  \label{fig:concept_examples}}
\end{figure}

\section{Introduction}
\input{intro.tex}

\section{Methods}\label{sec:method}
\input{methods.tex}

\section{Experiments}\label{sec:experiments}
\input{experiments.tex}

\section{Related Works}
\input{related_works.tex}

\section{Conclusion}
\input{conclusion.tex}

\section*{Reproducibility statement}
Our results can be reproduced, following the method described in~\autoref{sec:method} and~\autoref{sec:imp_details}. Corresponding code is provided as supplemental material.

\bibliography{iclr2026_conference}
\bibliographystyle{iclr2026_conference}

\clearpage
\appendix

\input{appendix.tex}

\end{document}

%% file: math_commands.tex
\usepackage{amsmath,amsfonts,bm}

\def\figref#1{figure~\ref{#1}}

\def\eqref#1{equation~\ref{#1}}

\def\1{\bm{1}}

\DeclareMathAlphabet{\mathsfit}{\encodingdefault}{\sfdefault}{m}{sl}
\SetMathAlphabet{\mathsfit}{bold}{\encodingdefault}{\sfdefault}{bx}{n}



%% file: intro.tex
Interpretability of neural network representations is essential for building trustworthy models, enabling a deeper understanding of the mechanisms underlying a model’s predictions, and promoting fairness and accountability.
However, interpreting the internal representations learned by neural networks remains a central challenge in deep learning. Sparse autoencoders (SAEs)~\cite{bricken2023towards, cunningham2023sparse} have emerged as a powerful tool for extracting sparse and semantically meaningful features from model activations. Nevertheless, they face challenges that limit their applicability. Notably, they suffer from difficulties in training, and may still yield polysemantic features, not corresponding to a single interpretable concept. Moreover, sparse autoencoders (and similar methods) rely on feature sparsity as a proxy for interpretability, a choice that has been criticized as potentially inadequate~\cite{sharkey2025open}.

We introduce an alternative to sparse autoencoders (SAEs) for extracting features that correspond to interpretable concepts from neural networks. Drawing inspiration from Deleuze’s philosophical view of concepts as differences, we model concepts as directions that capture distinctions between representations of individual samples.
Specifically, our approach can be seen as an unsupervised discriminant analysis: it identifies directions in the internal representation that best separate data samples.
We estimate those directions by sampling activation differences between pairs of data points, then use KMeans clustering to uncover recurring patterns.
Our analysis is further refined using distributional skewness to promote diversity.

Evaluating interpretability methods remains a major challenge. SAEs are often assessed by their reconstruction–sparsity trade-off, which does not necessarily reflect interpretability.
Hence, most recent studies in this field are also evaluated qualitatively, showing their relevance through selected examples. While insightful, such evaluations provide limited support. 
In contrast, we adopt a quantitative evaluation based on probe loss~\cite{gao2024scaling}, which measures the extent to which extracted concepts capture the attributes expected to be present in a dataset.
To ensure robust evaluation, we apply this metric to a broad set of 874 attributes spanning different tasks, five datasets and five models across three modalities (image, text and audio).
Our method captures the desired attributes more effectively than recent SAE-based approaches. In several settings, it is competitive with supervised linear discriminant analysis.
Beyond the presence of expected attributes, we also evaluate
cross-run consistency with the Maximum Pairwise Pearson Correlation (MPPC)~\citep{wang2025towards}, establishing a comprehensive evaluation framework for concept evaluation methods.
Finally, we demonstrate concept steering on text and image models, showing that manipulating extracted concepts causally influence downstream behavior, without incurring information loss.

Hence, the main contribution of this paper is a novel type of approach of mechanistic interpretability of neural networks. We investigate the fundamental principle underlying our approach and demonstrate that it achieves globally more compelling results than state-of-the-art sparse autoencoder (SAE)–based techniques. Our method is advantageous in its simplicity: it is governed by a single, interpretable hyperparameter. The proposed principle is theoretically grounded in discriminant analysis and clustering, and further relates to Deleuze’s philosophical notion of “concepts.” Similar to SAE-based approaches, our method is fully unsupervised and therefore does not require manual specification or annotation of the identified concepts.
Code is publicly available on GitHub\footnote{\url{https://github.com/ClementCornet/Deleuzian-Hypothesis}}.

%% file: methods.tex
\subsection{Criteria and Conceptual Grounding}\label{sec:criteria}

Our aim is to extract an ontology of ``concepts'' from a neural network, by analyzing its activations.
Before proposing our approach, we first discuss the criteria such
concepts should satisfy.

\begin{itemize}
    \item \textit{Interpretability}: this work aims to extract human-interpretable features, that are then referred to as ``concepts''.
    \item \textit{Transparency}: in order to gain interpretable insights into the model, the approach itself should be as simple and transparent as possible, not relying on non-interpretable hyperparameters.
    \item \textit{Diversity}: the extracted concepts should be semantically diverse, in order to represent a wide variety of data samples, ideas, and semantic levels.
    \item \textit{Consistency}: 
    the approach should consistently yield similar concepts when run multiple times with different random seeds.
\end{itemize}

Existing methods in mechanistic interpretability typically extract
unsupervised concepts by reconstructing model activations~\cite{bricken2023towards,cunningham2023sparse}. Because they are trained to minimize reconstruction error, such approaches are driven to capture as much variance in the activation space as possible, subject to sparsity constraints. This framing implicitly presents concepts as universal structural components of the model activations, echoing the classical philosophical view of concepts as ``the universal essence of a fact'' ~\cite{plato375republic_vi, hegel1816wissenschaft}. However, such a representation has been criticized as overly restrictive~\cite{nietzsche1889gotzen, sartre1946existentialisme}.
More recent perspectives instead emphasize concepts as arising from 
\textit{Difference and Repetition}~\cite{deleuze1968différence}, rather than universals.
Following this idea, our approach does not attempt to model the full variance
of activations. Instead, it identifies recurring differences between
activations.

\subsection{Extracting Repeated Differences in Activation Space}

\begin{figure}[htbp]
    \centering
    \includegraphics[width=\textwidth]{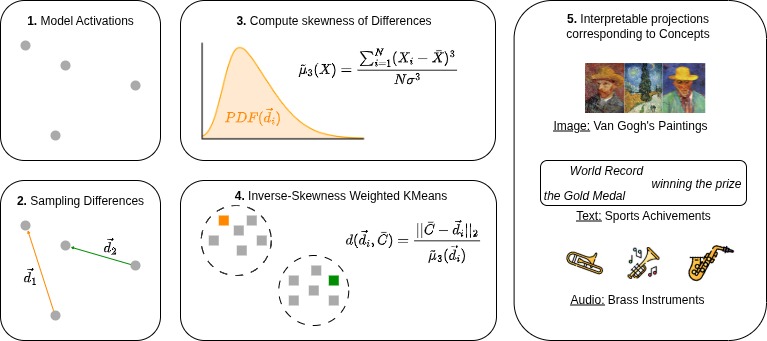}
    \caption{Overview of our concept extraction approach. We sample pairwise differences in activation between samples. Then, we use the inverse-skewness of those differences to selected the final concepts, corresponding to vectors in the activation space.}\label{fig:pipeline}
\end{figure}

Our objective is to extract concepts from model activations, at a given layer with $\mathcal{D}$ dimensions, over a dataset of $N$ samples.
To represent repeated differences in activations between data samples, we define $D = \{\vec{d_1}, \vec{d_2}, ..., \vec{d_N}\}$ as a set of $\mathcal{D}$-dimensional pairwise differences in activation between samples.
Since our approach is fully unsupervised, we cannot restrain $D$ to
contrastive pairs between two classes.
However, computing all pairwise differences is quadratic in $N$.
To approximate the distribution of differences, we instead randomly
sample $N$ pairs, ensuring that each data point is used once on each side of the subtraction.

To constrain our concept dictionary to a fixed number of concepts $k$, we cluster activation differences using KMeans~\cite{lloyd1982least, zeng2019cs}.
However, some activation differences exhibit highly skewed distributions: they remain near-zero for most samples, but occasionally spike to large values. Those differences tend to dominate the Euclidean distance used by standard KMeans, and produce redundant clusters~\cite{milligan1980examination}. 
The skewness of a distribution $X$, defined as the normalized third central moment is 
\begin{equation}
    \tilde{\mu}_3(X) = \frac{\sum^{N}_{i=1} (X_i - \bar{X})^{3}}{N \sigma^{3}}
\end{equation}

For a concept direction $\vec{d_i}$, we consider skewness as that of the projection $\{\vec{d_i} \cdot \vec{x}_j\}_{j=1}^N$. Since highly skewed coordinates tend to produce redundant clusters, we penalize them by assigning weights inversely proportional to skewness. 
In order to avoid ill-defined clustering with negative weights, and to consider opposite directions $\vec{d_i}$ as similar (as we are seeking directions, regardless of their orientation), we consider $-\vec{d_i}$ for differences with negative skewness.
This results in a variant of Feature-Weighted KMeans~\citep{huang2005automated}, in which concept directions are weighted during centroids computation, in order to promote concept diversity. More precisely, this clustering defines the weighted distance between $\vec{d_i}$ and its corresponding centroid $\bar{C}$ as 
$$
d(\vec{d_i}, \bar{C}) = \frac{1}{\tilde{\mu}_3(\vec{d_i})}||\bar{C} - \vec{d_i}||_2
$$
The obtained centroids are then used as concept vectors.
%

Both pair sampling and KMeans clustering run in linear time and memory with respect to dataset size $N$ and activation dimension $\mathcal{D}$, demonstrating scalability of our approach towards large datasets, or large models.

Finally, this procedure retains a simple and transparent formulation (~\autoref{fig:pipeline}), that are key properties for interpretability research. Notably, the number of extracted concepts $k$ is the only hyperparameter required for our approach, and is itself interpretable.

\subsection{Connection to Discriminant Analysis}\label{sec:lda}

{\color{black}
We aim to extract ``concepts'' from model activations, defining a concept as a difference between ideas. In a supervised setting, this objective relates closely to discriminant analysis~\cite{fisher36lda}, which identifies a direction $\vec{c}$ orthogonal to the optimal separating hyperplane between two classes.} Let $\Sigma_A$ and $\Sigma_B$ be the class covariances, and $\mu_A$ and $\mu_B$ their means. The separation between classes is maximized by:

\begin{equation}
    \vec{c} \propto (\Sigma_A + \Sigma_B)^{-1} (\vec{\mu}_A - \vec{\mu}_B)
    \label{eq:fischer-w}
\end{equation}

Consider two samples $i$ and $j$ with {\color{black}activations} $\vec{x}_i$ and $\vec{x}_j$, {\color{black}and} suppose we seek the optimal separation between clusters with means $\vec{x}_i$ and $\vec{x}_j$, distinguished by a concept $\vec{c}$. {\color{black}In high-dimensional spaces (typically $\ge 512$ dimensions for transformers), we approximate $\Sigma_i$ and $\Sigma_j$ as diagonal, containing each dimension's variance~\cite{ahdesmaki2010feature}. }
{\color{black}

From \eqref{eq:fischer-w}, $\vec{c} \propto \vec{x}_i - \vec{x}_j$ achieves optimal separation when $\Sigma_i \propto \Sigma_j \propto I$, i.e., under isotropic cluster distributions. Thus, treating activation differences as the optimal separation between ideas is equivalent to assuming isotropic distributions of concepts in activation space.
}

Unlike standard LDA, \eqref{eq:fischer-w} does not require homoscedasticity or Gaussianity~\cite{mclachlan2005discriminant}, and naturally extends to multiclass discrimination~\cite{rao1948utilization}.

{\color{black} In \autoref{app:quadratic_extension}, we derive a quadratic extension to our approach, that accounts for anisotropic distribution of concepts. While theoretically interesting, it does not lead to better experimental results. For this reason, we focus on the isotropic approach (\textit{i.e} $\Sigma_i \propto \Sigma_j \propto I$) in the following.}

\subsection{Lossless Steering}
\label{sec:steering}
Sparse autoencoders and related methods allow steering of extracted concepts~\cite{zhou2025llm}. To do so, they project sample activations in their concept space, apply a steering vector, and projects back into the activation space.
The two projections required introduce reconstruction error and information loss.
In contrast, our extracted concepts are vectors in the activation space. Therefore, we can perform steering directly in the
activations space.
To steer the embedding of a sample $x$, with a magnitude $\alpha$ and a concept $\vec{c_i}$, consider its steered representation $\tilde{x} = x + \alpha \vec{c_i}$.
{\color{black}If one steers a concept by $+\alpha$, then by $-\alpha$, we retrieve exactly the base activation.}
By avoiding projections into and out of the concept space, our approach enables lossless steering: the modifications affect only the targeted direction and can be exactly reversed.

%% file: experiments.tex

\paragraph{Datasets and Models}
To evaluate our concept extraction methods, we
conduct a large-scale study spanning five models and five
datasets across three modalities (vision, language, and
audio), covering a wide variety of semantic attributes.

For text, we use two datasets:
IMDB~\citep{maas-EtAl:2011:ACL-HLT2011} and
CoNLL-2003~\citep{tjong-kim-sang-de-meulder-2003-introduction}.
IMDB provides sentence-level binary sentiment
classification labels, while CoNLL-2003 provides
token-level labels for named entity recognition (NER),
part-of-speech (POS) tagging, and syntactic chunking.
For vision, we use a subset of ImageNet~\citep{imagenet15russakovsky} 
with 100 classes and the WikiArt dataset~\cite{wikiart}
which contains paintings labeled by artist (129 classes), style (27 classes), and genre (11 classes). Concerning text datasets, IMDB has binary classification labels, while CoNLL-2003 has token-wise labels for NER (9 classes), POS-tagging (47 classes) and chunk tags (23 classes).
For audio, we use AudioSet~\citep{jort_audioset_2017}, with multi-classification labels (527 audio classes).

Our text experiments are conducted on DeBERTa~\citep{he2021deberta} and the
encoder of BART~\citep{lewis2020bart}, {\color{black} as well as Pythia-70M~\cite{biderman2023pythia}}. For vision, we evaluate
DinoV2~\citep{oquab2023dinov2} and CLIP~\citep{radford2021learning}. For audio,
we use a pretrained Audio Spectrogram Transformer (AST)~\citep{gong2021ast}.
We only consider encoder models, (including the encoder of BART). This choice
allows us to evaluate the quality of extracted concepts with respect to
supervised labels that are likely represented at the analyzed layer of each
model, since our objective is to compare concept extraction methods. It also
enables comparable analyses across multiple modalities.
More details on datasets and models are provided
in~\autoref{app:exp_setup}.

\paragraph{Baselines}

Sparse autoencoders (SAE) are predominant among concept
extraction methods. We compare our method to five
different types of SAEs:

\newcommand{\VanillaSAE}{Van-SAE}
\newcommand{\GatedSAE}{Gat-SAE}
\newcommand{\JumpReLUSAE}{JR-SAE}
\newcommand{\MatryoshkaSAE}{Mat-SAE}
\newcommand{\TopKSAE}{Tk-SAE}
\newcommand{\ArchetypalSAE}{A-SAE}
\newcommand{\PretrainedSAE}{Pretrained}

\begin{itemize}
    \item VanillaSAE (\textbf{\VanillaSAE})~\citep{bricken2023towards}: standard SAE, trained with an $L_2$ reconstruction loss, and enforcing sparsity via an $L_1$ penalty which requires a coefficient $\lambda $;
    
    \item GatedSAE (\textbf{\GatedSAE})~\citep{rajamanoharan2024improving}: SAE learning activations gates, hence separating feature selection and magnitude estimation;

    \item  JumpReLUSAE (\textbf{\JumpReLUSAE})~\citep{rajamanoharan2024jumping}: SAE with a learnable threshold $\theta_i$ for each concept, designed to minimize the reconstruction error;
    
    \item MatryoshkaSAE (\textbf{\MatryoshkaSAE})~\citep{bussmannlearning}: SAE learning nested dictionaries of concepts, focusing on
    hierarchies of concepts, belonging to multiple semantic levels;
    
    \item TopKSAE (\textbf{\TopKSAE})~\citep{gao2024scaling}: SAE enforcing sparsity via a
    \textit{TopK} activation function, that sets every activation to zero,
    except the $k$ highest.
    
    {
    \color{black}
    \item ArchetypalSAE (\textbf{\ArchetypalSAE})~\cite{felarchetypal}: SAE constraining decoder atoms to be combinations of activations, to gain stability.
    
    \item Pretrained Sparse Autoencoders (\textbf{\PretrainedSAE}): we compare our method with publicly available, pretrained sparse autoencoders on two models. For DinoV2 experiments, we use ViT-Prisma~\cite{joseph2025prisma}, and for Pythia we use a sparse autoencoder trained by EleutherAI\footnote{\url{https://huggingface.co/EleutherAI/sae-pythia-70m-32k}}. 
    }

\end{itemize}
We also compare our approach to Independant Component Analysis (\textbf{ICA})~\citep{comon1994independent}, that is a linear decomposition method maximizing statistical independence between latent dimensions.
In addition, as our approach is closely related to discriminant analysis, we also compare it to supervised Linear Discriminant Analysis (\textbf{LDA})~\citep{fisher36lda} which serves as an upper bound under assumptions of homoscedasticity and normal distribution of concepts.

\paragraph{Evaluation}

Our primary quantitative evaluation relies on the probe loss metric~\cite{gao2024scaling}, which measures the degree to which extracted concepts align with ground-truth annotated attributes. Beyond the quality on individual concepts, we also aim at uncovering a broad set of concepts from model activations. 
To this end, we assess probe loss across tasks characterized by diverse attribute sets, thereby quantifying the capacity of our approach to capture multiple, semantically meaningful concepts.
In addition, Maximum Pairwise Pearson Correlation (MPPC)~\citep{wang2025towards} is used to measure the consistency of the different methods. 
Finally, to highlight causal influence of concepts on model predictions, we perform concept steering, and provide qualitative examples. 
Note that, while prior work on sparse autoencoders has emphasized reconstruction–sparsity trade-offs, these objectives are not applicable to our framework; we therefore exclude them from evaluation. 
All the reported results are computed using activations from the last transformer block of each encoder, using a concept space with 6144 dimensions, corresponding to 8 times the size of the activations (except for ICA, that is limited to 768).

\subsection{Evaluation of Concept Quality}


We evaluate concepts extracted in an unsupervised manner {\color{black}by assessing whether they correspond to interpretable attributes known to exist in the dataset. This correspondence is quantified using Probe Loss~\citep{gao2024scaling}.
For each attribute, Probe Loss measures how well a one-dimensional logistic probe can recover the ground-truth attribute from the extracted concepts. Specifically, we train a separate 1D logistic probe for every concept and record the lowest cross-entropy loss achieved. For multi-class attributes, we report the median Probe Loss across all attributes. The results of this evaluation are presented in \autoref{tab:probe-loss-all}.
}


\begin{table*}[t]
    \centering
    \caption{Quantitative evaluation (Probe Loss, lower is better) of unsupervised approaches on CLIP and DinoV2 image encoders, DeBERTa and BART text encoders and Audio Spectrogram Transformer on audio. Supervised baseline (LDA) is reported for reference (gray row). Best results are in \textbf{bold}, second in \textit{italics}. Bottom right table indicates the average rank of all methods over all datasets (lower is better). {\color{black}``Pretrained'' are models independently trained by other teams (see text for details)}}\label{tab:probe-loss-all}    
    \begin{tabular}{@{}llcccccccc@{}}
    \toprule
     & & \multicolumn{4}{c}{CLIP} & \multicolumn{4}{c}{DinoV2} \\
    \cmidrule(lr{0.1em}){3-6} \cmidrule(lr{0.1em}){7-10}
    \multirow{3}{*}{\rotatebox{90}{labels}}
     & \multirow{2}{*}{Method}
     & \multirow{2}{*}{ImNet}
     & \multicolumn{3}{c}{WikiArt}
     & \multirow{2}{*}{ImNet}
     & \multicolumn{3}{c}{WikiArt} \\
    \cmidrule(lr{0.1em}){4-6} \cmidrule(lr{0.1em}){8-10}
     & & & Artist & Style & Genre & & Artist & Style & Genre \\
    \midrule
    \rowcolor{gray!20}
    \ding{51} & LDA & 0.0083 & 0.0084 & 0.0465 & 0.0976 & 0.0044 & 0.0101 & 0.0545 & 0.1084 \\
    \midrule
    \ding{55} & ICA & \textit{0.0154} & 0.0141 & 0.0816  & 0.2104 & 0.0161 & 0.0155 & 0.0839 & 0.2035 \\
    \ding{55} & \VanillaSAE & 0.0264 & 0.0137 & \textbf{0.0558} & 0.1531 & 0.0220 & 0.0147 & 0.0722 & 0.1706 \\
    \ding{55} & \GatedSAE & 0.0384 & 0.0142 & 0.0747 & 0.1647 & 0.0345 & 0.0151 & 0.0789 & 0.1752 \\
    \ding{55} & \JumpReLUSAE & 0.0355 & 0.0138 & 0.0667 & 0.1490 & 0.0327 & 0.0148 & 0.0741 & 0.1723 \\
    \ding{55} & \MatryoshkaSAE & 0.0216 & 0.0141 & 0.0686 & 0.1588 & 0.0127 & 0.0154 & 0.0767 & 0.1613 \\
    \ding{55} & \TopKSAE & \textit{0.0154} & \textit{0.0125} & \textbf{0.0558} & \textit{0.1360} & \textit{0.0096} & \textit{0.0144} & 0.0718 & 0.1577 \\
    
    \ding{55} & {\color{black}\ArchetypalSAE} & \color{black}0.0172 & \color{black}0.0130 & \color{black}0.0567 & \color{black}0.1370 & \color{black} \textit{0.0143} & \color{black}0.0145 & \color{black}\textit{0.0713} & \color{black}\textbf{0.1429}\\
    \ding{55} & {\color{black}\PretrainedSAE} & - & - & - & - & \color{black}0.0333 & \color{black}0.0149 & \color{black}0.0787 & \color{black}0.1796 \\
    
    \ding{55} & Deleuzian (Ours) & \textbf{0.0128} & \textbf{0.0119} & \textit{0.0560} & \textbf{0.1230} & \textbf{0.0055} & \textbf{0.0137} & \textbf{0.0680} & \textit{0.1538} \\
    \bottomrule
    \end{tabular}

    
    
    \begin{tabular}{@{}llcccccccc@{}}
    \toprule
     & & \multicolumn{4}{c}{DeBERTa} & \multicolumn{4}{c}{BART} \\
    \cmidrule(lr{0.1em}){3-6} \cmidrule(lr{0.1em}){7-10}
    \multirow{3}{*}{\rotatebox{90}{labels}}
     & \multirow{2}{*}{Method}
     & \multirow{2}{*}{IMDB}
     & \multicolumn{3}{c}{CoNLL-2003}
     & \multirow{2}{*}{IMDB}
     & \multicolumn{3}{c}{CoNLL-2003} \\
    \cmidrule(lr{0.1em}){4-6} \cmidrule(lr{0.1em}){8-10}
     & & & NER & POS & Chunk & & NER & POS & Chunk \\
    \midrule
    \rowcolor{gray!20}
    \ding{51} & LDA & 0.6394 & 0.0429 & 0.0044 & 0.0062 & 0.3473 & 0.6326 & 0.3875 & 0.0870 \\
    \midrule
    \ding{55} & ICA & 0.6936 & 0.1251 & 0.0195 & \textit{0.0126} & 0.6931 & 1.4578 & 0.7143 & 6.1319 \\
    \ding{55} & \VanillaSAE & 0.6893 & 0.0869 & 0.0252 & 0.0173 & 0.5983 & \textit{0.2719} & \textit{0.1647} & 0.0447 \\
    \ding{55} & \GatedSAE & 0.6883 & 0.1223 & 0.0251 & 0.3982 & 0.6391 & 0.3982 & 0.4054 & 0.3208 \\
    \ding{55} & \JumpReLUSAE & 0.6908 & 0.1150 & 0.0248 & 0.0170 & 0.6931 & 0.4416 & 0.2111 & 0.0883  \\
    \ding{55} & \MatryoshkaSAE & \textbf{0.6836} & 0.0868 & 0.0189 & 0.0164 & 0.6931 & 1.120 & 0.4954 & 0.2143 \\
    \ding{55} & \TopKSAE & 0.6858 & 0.0839 & 0.0166 & 0.0167 & 0.5980 & 0.3478 & 0.2045 & \textbf{0.0399} \\
    \ding{55} & {\color{black}\ArchetypalSAE} & \color{black} 0.6859 & \color{black}\textit{0.0775} & \color{black}\textbf{0.0141} &\color{black} \textbf{0.0058} & \color{black}\textbf{0.5547} & \color{black}0.3754 & \color{black}0.1959 & \color{black}\textit{0.0415} \\
    \ding{55} & Deleuzian (Ours) & \textit{0.6849} & \textbf{0.0665} & \textit{0.0161} & 0.0143 & \textit{0.5974} & \textbf{0.2148} & \textbf{0.0639} & 0.0419 \\
    \bottomrule
    \end{tabular}


    \begin{tabular}{@{}llcccccc@{}}
    \toprule
     & & AST & \multicolumn{3}{c}{\color{black} Pythia} & \multicolumn{1}{c}{Avg. Rank  $\downarrow$ } \\
    \cmidrule(lr{0.1em}){3-3} \cmidrule(lr{0.1em}){4-6} \cmidrule(lr{0.1em}){7-7}
    \multirow{3}{*}{\rotatebox{90}{labels}} 
     & \multirow{2}{*}{Method} 
     & AudioSet 
     & \multicolumn{3}{c}{\color{black}CoNLL-2003}
     & \multirow{2}{*}{ } \\
    \cmidrule(lr{0.1em}){4-6}
    & & & {\color{black}NER} & {\color{black}POS} & {\color{black}Chunk} & \\
    \midrule
    \rowcolor{gray!20}
    \ding{51} & LDA & 0.0164 & {\color{black}0.0742} & 0.0072 & 0.0089 & - \\
    \midrule
    \ding{55} & ICA & 0.0234 & {\color{black}0.1378} &\color{black} 0.0331 & \color{black}0.0088 & \color{black}6.85$\pm$2.29 \\
    \ding{55} & \VanillaSAE & 0.0177 & {\color{black}0.1498} & {\color{black}0.0272} & \color{black} 0.0083 & \color{black} 4.65$\pm$1.56 \\
    \ding{55} & \GatedSAE & 0.0186 & \color{black}0.1480 & \color{black} 0.0231 & {\color{black}0.0086} & \color{black} 6.65$\pm$1.42 \\
    \ding{55} & \JumpReLUSAE & 0.0181 & \color{black}0.1507 & \color{black}0.0277 & {\color{black}0.0085} &  \color{black}5.75$\pm$0.94\\
    \ding{55} & \MatryoshkaSAE & 0.0186 & \color{black}0.1754 & \color{black}0.0320 & {\color{black}0.0088} &\color{black} 5.70$\pm$1.90 \\
    \ding{55} & \TopKSAE & \textit{0.0169} & \color{black}\textit{0.1321} &\color{black} \textit{0.0203} & \color{black}\textit{0.0082} & \color{black} \textit{2.65$\pm$1.01} \\
    \ding{55} & {\color{black}\ArchetypalSAE} & \color{black} \textit{0.0169} & \color{black} 0.1378 & \color{black}0.0331 & \color{black}0.0088 & {\color{black}3.20$\pm$1.72} \\
    \ding{55} & {\color{black}\PretrainedSAE} & - & {\color{black}0.1717} & {\color{black}0.0344} & {\color{black}0.0087} &  - \\ 
    \ding{55} & Deleuzian (Ours) & \textbf{0.0164} & \textbf{\color{black}0.1121} & \textbf{\color{black}0.0133} & \textbf{\color{black}0.0080} & \color{black} \textbf{1.65$\pm$0.85} \\
    \bottomrule
\end{tabular}

\end{table*}

From~\autoref{tab:probe-loss-all},
our method globally outperforms all variations of SAE, with the lowest probe loss on 13 of the 20 tested tasks. This indicates a high ability to recover attributes expected to be found in datasets, on a wide variety of tasks, models and modalities.
On several cases, probe loss is midway between supervised LDA and the second most effective unsupervised method (typically TopKSAE).
Note that LDA obtains poor results on BART over CoNLL-2003, which indicates that the additional hypothesis made by LDA compared to our method (normal distribution of concepts and homoscedasticity) are not satisfied in this particular case. On average over all datasets, our approach is significantly the best classified among unsupervised approaches. Significance of the results is detailed in~\autoref{sec:supp:signif_probloss}.
    
To complement the quantitative evaluation, we further analyze representative examples, which provide evidence for the relevance and interpretability of the extracted concepts: in addition to the examples provided in~\autoref{fig:concept_examples}, we present qualitative results in~\autoref{app:concepts}.

\subsection{Consistency Across Runs}

%
%
%

In order to measure consistency of a concept extraction method, we measure the Maximum Pairwise Pearson Correlation (MPPC)~\citep{wang2025towards} 10 times between sets of concepts extracted with different random seeds, and report the average. Therefore, a MPPC closer to 1 indicates a higher consistency. 
We present MPPC in details and discuss its statistical significance in~\autoref{app:mppc}.

\begin{table}[tb]
\centering
\caption{Evaluating the consistency of extracted concepts with MPPC on several tasks/datasets including WikiArt (WA), AudioSet (AS).}
\label{tab:mppc}
\begin{tabular}{@{}llllllllll@{}}
\toprule
& \multicolumn{2}{c}{CLIP} & \multicolumn{2}{c}{DinoV2} & \multicolumn{2}{c}{DeBERTa} & \multicolumn{2}{c}{BART} & \multicolumn{1}{c}{AST} \\ \midrule
& ImNet & WA & ImNet & WA & IMDB & CoNLL & IMDB & CoNLL & AS \\
\midrule
ICA             & 0.449 & 0.388 & 0.264 & 0.406 & 0.122 & 0.440 & \textit{0.999} & 0.420 & 0.296 \\
\VanillaSAE      & \textbf{0.840} & \textbf{0.918} & \textit{0.603} & \textbf{0.903} & \textbf{0.986} & 0.437 & 0.996 & 0.439 & \textbf{0.837} \\
\GatedSAE        & 0.346 & 0.415 & 0.264 & 0.401 & 0.836 & 0.453 & 0.996 & 0.357 & 0.399 \\
\JumpReLUSAE     & 0.341 & 0.440 & 0.272 & 0.424 & 0.894 & 0.536 & 0.996 & 0.439 & 0.449 \\
\MatryoshkaSAE   & 0.225 & 0.247 & 0.201 & 0.219 & 0.707 & 0.339 & 0.506 & 0.216 & 0.274 \\
\TopKSAE         & 0.757 & \textit{0.861} & 0.588 & 0.824 & 0.866 & 0.594  & 0.996 & \textit{0.761} & 0.601 \\
Deleuzian (Ours)            & \textit{0.821} & 0.856 & \textbf{0.789} & \textit{0.843} & \textit{0.980} & \textbf{0.588} & \textbf{1.0} & \textbf{0.768} & \textit{0.830} \\
\bottomrule
\end{tabular}
\end{table}

Results from \autoref{tab:mppc} show that our approach generally extracts more consistent concepts than other models, except for VanillaSAE, but this method reaches much lower concept quality and diversity according to~\autoref{tab:probe-loss-all}.

\subsection{Concept Steering: Qualitative Evidence of Causal Influence}

A possible use of extracted concepts is to explicitly modify the behavior of a model, by steering its internal concepts. We provide qualitative examples of steering, using the method described in~\ref{sec:steering}, highlighting the causal influence of concepts on the output of a model.

\begin{figure}[htbp]
    \centering
    \includegraphics[width=0.9\textwidth]{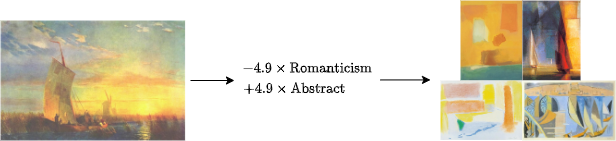} 
    \caption{Steering a painting style in CLIP activations: target is represented by its nearest images. \textit{Romanticism} is set to zero, while \textit{Abstract} is steered positively by the same magnitude.}\label{fig:steerclip}
\end{figure}

\paragraph{Discriminative Steering on CLIP}
{\color{black} Steering the inner representation of an image encoder may be used to perform style transfer, in a similar fashion as previous works~\cite{wynen2018unsupervised}.} From WikiArt, we consider two concepts corresponding to artistic styles (identified empirically from images), namely Romanticism and Abstract paintings. Starting from a romantic painting of a sailing ship, we inhibit the \textit{Romanticism} concept, and boost the \textit{Abstract paintings} one.  The resulting steered
embedding shifts the painting's representation such that its nearest neighbors in the WikiArt dataset are abstract sailing ships (\autoref{fig:steerclip}).

\paragraph{Steering BART}

\begin{figure}[htbp]
    \centering
    \includegraphics[width=\textwidth]{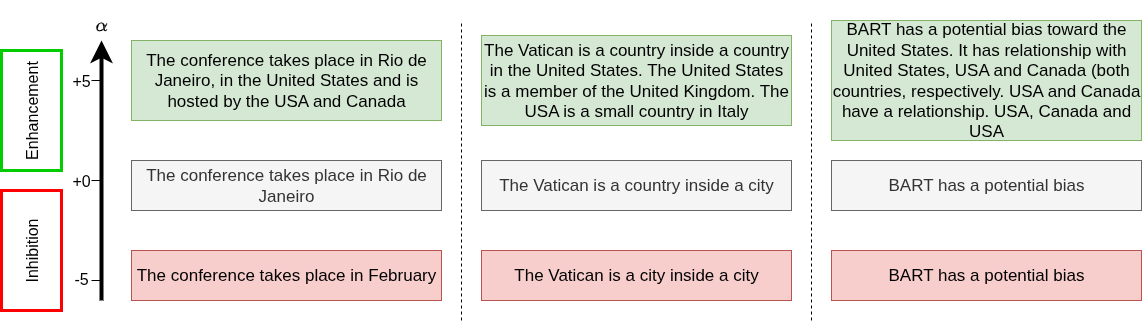}
    \caption{Steering the concept of \textit{countries} in a BART model for three sentences (in gray), using {\color{black} $\alpha=+5$ and $\alpha=-5$ in each case} }\label{fig:steerbart}
\end{figure}

BART~\citep{lewis2020bart} is a text encoder–decoder model which, without
finetuning, typically reproduces its input sequence. Here, we steer the
final transformer layer of its encoder before passing the modified
representation into the decoder. We analyze the steering effects of a
concept with highest activations corresponding to country names (\autoref{fig:steerbart}).\@
Inhibiting this concept ($\alpha < 0$) causes BART to replace ``Rio de Janeiro''
with ``February'', forming a coherent sentence with no geographical indication. In the same fashion, its leads to replacing the word ``country'' by the word ``city''.
Positive values of $\alpha$ encourage the model to evoke country names, even in sentences without geographic context. In particular, this leads to frequent mentions of the United States, highlighting a potential bias in BART.

\subsection{Ablation Studies}

\begin{table}[tbh]
    \centering
    \caption{Ablation study in terms of performance (Probe Loss) and diversity (effective rank). Our approach is the last line.}\label{tab:ablation}
    \begin{tabular}{@{}llc cccccc@{}}
    \toprule
     & & \multirow{3}{*}{\rotatebox{90}{\parbox{1.2cm}{\small skewness \\weighting}}} &\multicolumn{2}{c}{probe loss $\downarrow$} & \multicolumn{2}{c}{effective rank $\uparrow$ }  & \multicolumn{2}{c}{{\color{black}max. pairwise cos. $\downarrow$ }} \\
    input  & concept &   & CLIP  & DeBERTa & CLIP    & DeBERTa & {\color{black}CLIP}    & {\color{black}DeBERTa}    \\ 
    space & identif. &  & WikiArt  & CoNLL & WikiArt & CoNLL & {\color{black}WikiArt} & {\color{black}CoNLL}    \\ \midrule
    acts. & \TopKSAE  & \ding{55} &  \textit{0.0125}     & \textit{0.0839} & 96.1   & \textbf{183.9} & {\color{black}\textbf{0.2900}} & {\color{black}\textit{0.3716}}\\
    acts. & KMeans  & \ding{51} &   0.0133    & 0.1184 &   24.3  & 14.6 & {\color{black}0.8685} & {\color{black}0.9195} \\ 
    diff & \TopKSAE   & \ding{55} &   0.0134    & 0.1093 &  \textbf{340.5}  & 109.2 & {\color{black}\textit{0.3407}} & {\color{black}\textbf{0.1737}} \\
    diff &  KMeans   & \ding{55} &   0.0128    & 0.0841 &   17.9  & 5.65 & {\color{black}0.6504} & {\color{black}0.8357}  \\ \midrule
    diff & KMeans    & \ding{51} &   \textbf{0.0119}    & \textbf{0.0665} & \textit{124.4}   & \textit{182.0} & {\color{black}0.5677} & {\color{black} 0.3908} \\ \bottomrule
    \end{tabular}
\end{table}


We conduct an ablation study of our method, to assess the impact of three aspects on its performance. First, we evaluate the interest of learning from differences between samples, rather than directly from the samples themselves (i.e. changing the input space). Second, we evaluate the impact of using a clustering to identify the concepts, by replacing the the KMeans clustering of our approach with an SAE, trained on the activations or the differences. Finally, we evaluate the impact of weighting the KMeans clustering by the inverse skewness. Since the objective of this weighting is to increase diversity, we also report an evaluation of the diversity of the extracted concepts, measured by the effective rank \cite{roy2007effective,skeanlayer} {\color{black}, as well as the maximum pairwise cosine among concept directions that quantifies redundancy}.
Results, computed on CLIP activations on WikiArt, and DeBerta on CoNLL NER attributes, are reported in~\autoref{tab:ablation}. 
These results, most notably those for KMeans on activations and TopKSAE on differences highlight the impact of representing differences in activations. Moreover, these results highlight the importance of using the inverse skewness of pairwise differences as KMeans weights, allowing the extraction of a much larger, {\color{black} and less redundant sets of concepts, according to both effective rank and maximum pairwise cosine metrics}.

\begin{figure}[htbp]
    \centering
    \includegraphics[width=\textwidth]{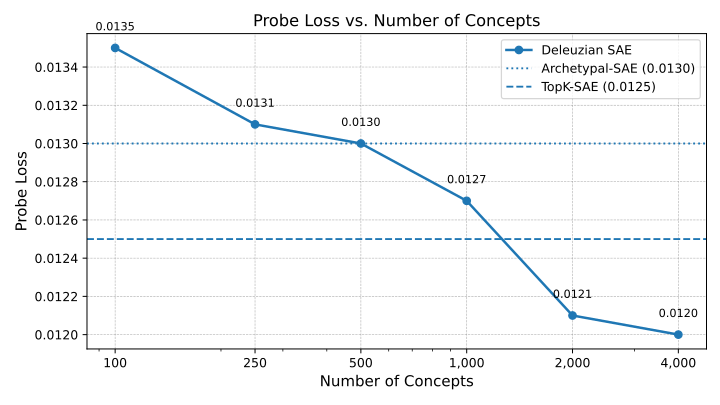}
    \caption{Performance of our Deleuzian approach using less than 6144 concepts, on CLIP, WikiArt artist task.}\label{fig:lessconcepts}
\end{figure}

\autoref{fig:lessconcepts} evaluates the performance of our method while extracting a number of concepts smaller than 6144. Only 2000 concepts are needed to outperform every concurrent method on CLIP, WikiArt artist task. This highlights the ability of our method to \textit{efficiently} recover concepts.

%% file: related_works.tex
\textbf{Concept-Based Interpretability}
Identifying the internal mechanism of a neural network corresponding to a
precise concept provides valuable insights into the network's behavior.
~\citet{arik2020explaining} perform clustering on multi-layer activations, in order to determine similar images, not to extract interpretable concepts.
Prior studies have investigated the extent to which a classification probe
can be learned directly on model hidden representations~\cite{kohn2015s}.
Probe-based concept extraction has been used extensively in
NLP~\cite{gupta2015distributional}. These studies suggest that LLMs linearly
represent the truth or falsehood of factual statements~\cite{marks2024the}.
Similar analyses have also been applied to computer vision~\cite{alain2017understanding} or reinforcement
learning~\cite{lovering2022evaluation}. However, probe-based concept
extraction only captures correlation (not causation) and heavily relies on
curated data to extract concepts~\cite{belinkov2022probing}.
To address this problem, Concept Bottleneck Models (CBM)~\cite{koh2020concept}
structure the network to make predictions through a layer of
human-defined concepts, enabling intervention but requiring labeled concept
supervision. 
Contrast-Consistent Search probes for an axis in the activation space,
corresponding to the presence or absence of a
concept~\cite{burnsdiscovering}, however it uses predefined contrastive groupings,
and thus cannot uncover new concepts. Similarly, TCAV~\cite{kim2018interpretability} and ACE~\cite{ghorbani2019towards} perform concept extraction upon a predefined list. 

\textbf{Sparse Autoencoders}
Sparse autoencoders (SAEs)~\cite{lee2007sparse} are a sparse dictionary
learning technique that aims to find a sparse decomposition of data into an
overcomplete set of features. They typically enforce sparsity via an $L_1$
penalty. In recent years, SAEs have been applied to neural networks to learn
an unsupervised dictionary of interpretable features tied to concepts from a
hidden representation~\cite{bricken2023towards, cunningham2023sparse}.
Various extensions of sparse autoencoders have been proposed with modified
activation functions, such as JumpReLU~\cite{rajamanoharan2024jumping},
TopK~\cite{gao2024scaling}, and BatchTopK~\cite{bussmann2024batchtopk} sparse
autoencoders. Other works seek hierarchies of features by extracting nested
dictionaries~\cite{bussmannlearning, zaigrajewinterpreting}. Analogous
methods have been developed in order to find relations between different
layers of a same network, including
transcoders~\cite{dunefsky2024transcoders} and
crosscoders~\cite{Lindsey2024SparseCrosscoders}.{\color{black} ArchetypalSAE~\cite{felarchetypal} constrains the decoder training in order to gain stability, while Spade~\cite{hindupur2025projecting} is a distance based SAE. }

\textbf{Further use of extracted concepts}
Identifying the mechanism corresponding to a semantic concept within a neural network enables new uses of the analyzed model. For example, studies use extracted concepts to analyze the circuits related to a specific task~\cite{conmy2023towards, dunefsky2024transcoders}, or to measure the importance of concepts in model inner representations~\cite{fel2023holistic}. Concept
extraction techniques can also be used to perform \textit{steering}, i.e.
controlling the behavior of a model by explicitly modifying its internal concepts~\cite{zhou2025llm}. When applied to multiple models in parallel, concept extraction methods allow construction of shared concept spaces~\cite{thasarathan2025universal}, automating naming of CLIP concepts ~\citep{rao2024discover} and quantification of similarities between models~\cite{wang2025towards}.

%% file: conclusion.tex
\paragraph{Discussion}
%
{\color{black} 
We present a novel approach for extracting human-interpretable ``concepts'' from neural network activations and evaluate it across five models and three modalities. Our method is simple and can be interpreted as an unsupervised form of discriminant analysis. Probe loss evaluation shows that the extracted concept space captures attributes expected from labeled datasets, and our approach outperforms existing methods on this metric. Moreover, the concepts are stable across multiple runs, enabling consistent analyses, and the method supports lossless interventions on internal representations. These results suggest that explicitly representing inter-sample \textit{differences}, in line with Deleuze’s notion of concepts, can improve both the quality and utility of extracted concepts.
}

\paragraph{Limitations}
%
{\color{black} 
Although our method is fully unsupervised, its evaluation depends on labeled datasets. Consequently, interpretable concepts that do not align with the available labels may incur high probe losses, even if they are highly meaningful but subtle or specific. Evaluating without labels would require a theoretically justified proxy for interpretability, which remains lacking; sparsity alone does not satisfy this criterion~\cite{sharkey2025open}.

All evaluations are performed in concept spaces of 6,144 dimensions (8× the activation dimension), except for an ablation. While some studies use even higher-dimensional projections, further increasing dimensionality could bias our evaluation, given the limited number of attributes and data samples relative to the potential size of the concept space. Exploring higher-dimensional spaces could nonetheless reveal additional characteristics of concept extraction methods.

Our approach assumes that concepts can be represented as linear projections. This assumption is empirically validated across five models spanning different categories and modalities. However, a model with inner representations that violate this assumption could exist and would require adapting the method.

}

\paragraph{Perspectives}

%
%
Our method is fully unsupervised and extracts concepts that represent repeated directions in a model. 
{\color{black}
Consequently, a method that can automatically name or interpret these concepts would greatly enhance the scope and applicability of the findings, enabling more comprehensive analyses across datasets, modalities, and models. Such generalization could facilitate understanding of model behavior, provide interpretable axes for interventions, and support downstream tasks that leverage concept-level information.
}
We provide qualitative examples of concept steering. As our method allows lossless steering, such intervention on model inner representations could be used at a larger scale, for example to adapt to a specific domain.


\noindent
\textbf{Acknowledgment} this work was partially funded by the Agence Nationale de la Recherche (ANR) for the STUDIES project  ANR-23-CE38-0014-02. It was made possible by the use of the FactoryIA supercomputer, financially supported by the Ile-De-France Regional Council.

%% file: appendix.tex
\section{Appendix: Implementation details}\label{sec:imp_details}

All our experiments are using a set of 6144 concepts, except for ICA, that is unable to represent a number of dimensions larger than $\mathcal{D}$, the dimension of model activations. Therefore, ICA experiments are ran in $\mathcal{D}=768$ dimensions. 

TopKSAEs are trained using a TopK activation function, with $k=32$. We select a learning rate of $10^{-5}$, that minimizes its reconstruction error on CLIP activations over ImageNet. For VanillaSAE, GatedSAE and JumpReLUSAE, we select the $L_1$ penalization coefficient reaching the lowest probe loss. From a sweep of 7 values between $10^{-9}$ and $10^{-3}$, we select $10^{-8}$ for VanillaSAE, $10^{-6}$ for GatedSAE and $10^{-5}$ for JumpReLUSAE. Concerning MatryoshkaSAE, we use groups of sizes  $[512, 1024, 1536, 3072]$, in order to represent progressively larger latent dictionaries. 

For Independent Component Analysis we used the scikit-learn~\cite{scikit-learn} implementation of FastICA~\cite{hyvarinen2000ica_algo_appli}, with a log hyperbolic cosine to approximate the neg-entropy, a SVD whitening and the extraction of multiple components in parallel.

\section{Appendix: Details on Experimental Setup}\label{app:exp_setup}
All datasets used in our experiments (\autoref{sec:experiments}) are reported in \autoref{tab:datasets} with their main characteristics. When available, we use the train/test splits provided. 
As WikiArt has no predefined train/test sets, we use its even samples (0, 2, 4...) as a train set, and the other ones as the test set. Note that WikiArt is actually a set of data with three different label types, thus could be considered as three different datasets.

Globally we thus have a much larger variety of experimental settings than in comparable previous works. Since we are interested in identifying concepts, all tasks relate to classification but they exhibit a deep variety in their nature, due to the type of data handled (text, image, audio) and how the data have to be considered to address the task. For example, the identifying \textit{sentiments} on IMDB requires to take into account full sentences while the \textit{chunking} task in CoNLL act at the token level.

\begin{table}[ht]
\centering
\small
\caption{Datasets used in our experiments.}\label{tab:datasets}
\begin{tabular}{l l l l l }
\toprule
\textbf{Dataset} & \textbf{Modality} & \textbf{Label Type (number of classes)} & \textbf{Train/Test Size} & URL\\
\midrule
ImageNet-100 & Image & Object categories (100) & 50k / 5k & \href{https://huggingface.co/datasets/timm/mini-imagenet}{\faDownload} \\
WikiArt & Image & Artist (129), Style (27), Genre (11) & 40k / 40k & \href{https://huggingface.co/datasets/huggan/wikiart}{\faDownload}\\
IMDB & Text & Sentiment (binary, sentence-level) & 25k / 25k & \href{https://huggingface.co/datasets/stanfordnlp/imdb}{\faDownload}\\
CoNLL-2003 & Text & NER (9), POS (47), Chunking (23, token-level) & 288k / 67k & \href{https://huggingface.co/datasets/eriktks/conll2003}{\faDownload}  \\
AudioSet & Audio & Audio event categories (527) & 18k / 17k & \href{https://huggingface.co/datasets/agkphysics/AudioSet}{\faDownload} \\
\bottomrule
\end{tabular}
\end{table}

The model encoders we considered in our experiments are summarized in \autoref{tab:models}.  All the models were downloaded from huggingface, except for CLIP from OpenClip ~\cite{ilharco_gabriel_2021_5143773} and DinoV2 from PyTorch Hub. The \textit{model size} is the number of parameters and since all of them were encoded in \texttt{float32} their actual size in memory is this number multiplied by four. 

AST~\cite{gong2021ast} relies on an image ViT that was trained on ImageNet-21k then finetuned on AudioSet. 
BART~\cite{lewis2020bart}, for its \textit{base} version, was pre-trained  ``on the same data as BERT~\cite{devlin-etal-2019-bert}'' that is ``a combination of books and Wikipedia data''. 
CLIP~\cite{radford2021learning} was trained ``on publicly available image-caption data'' that is images-caption pairs from the Web and publicly available datasets such as YFCC 100M~\cite{thomee2016yfcc}. The creator of the model did not release the dataset to avoid its use ``as the basis for any commercial or deployed model''.
DeBERTa~\cite{he2021deberta} was trained on deduplicated data (78G) including original Wikipedia (English Wikipedia dump; 12GB), BookCorpus  (6GB), OpenWebText (public Reddit content; 38GB), and STORIES (a subset of CommonCrawl; 31GB). 
DinoV2~\cite{oquab2023dinov2} was trained on the LVD-142M dataset, that was assembled and curated by the authors of the model.

\begin{table}[ht]
\centering
\small
\caption{Pretrained models used in our experiments. The \textit{Size} is the number of parameters (in millions).}\label{tab:models}
\begin{tabular}{l l l r l c}
\toprule
\textbf{Model} & \textbf{Modality} & \textbf{Version} & \textbf{Size} & \textbf{Training data} & \textbf{URL}\\
\midrule
DeBERTa & Text & base & 99 M & \parbox{5cm}{BookCorpus, Wikipedia, OpenWebText, STORIES} & \href{https://huggingface.co/microsoft/deberta-base}{\faDownload} \\
BART (encoder) & Text & base& 139 M & Books, Wikipedia& \href{https://huggingface.co/facebook/bart-base}{\faDownload} \\
DinoV2 & Image & ViT-B/14 & 86 M& LVD-142 & \href{https://dl.fbaipublicfiles.com/dinov2/dinov2_vitb14/dinov2_vitb14_pretrain.pth}{\faDownload}\\
CLIP & Image & ViT-B/16 & 150 M & openAI private: web, YFCC100M... & \href{https://github.com/mlfoundations/open_clip}{\faDownload} \\
AST & Audio & 10-10-0.4593  & 87 M & AudioSet, ImageNet-21k & \href{https://huggingface.co/MIT/ast-finetuned-audioset-10-10-0.4593}{\faDownload} \\
\bottomrule
\end{tabular}
\end{table}

\section{Appendix: Significance of Probe Loss Results}\label{sec:supp:signif_probloss}

\begin{figure}
    \centering
    \includegraphics[width=0.8\textwidth]{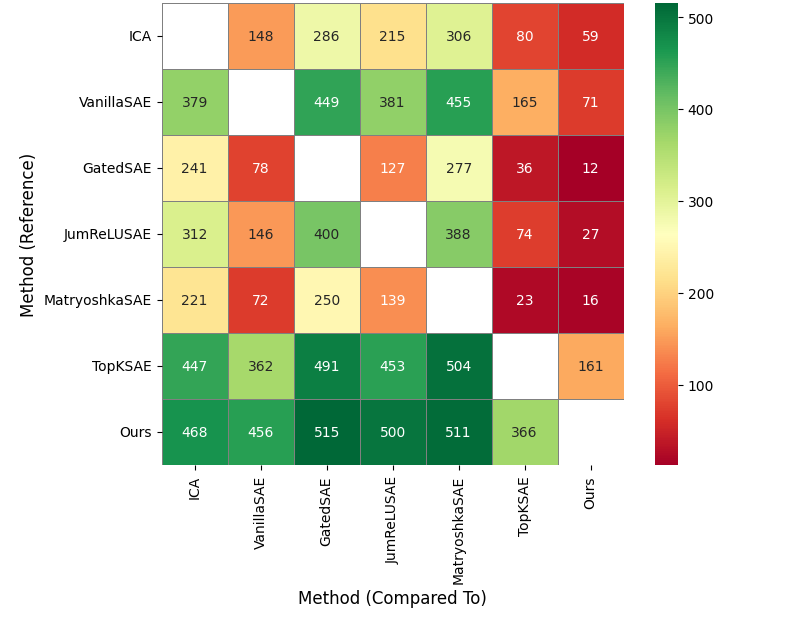}
    \caption{Pairwise comparisons of methods on AST-Audioset. Our method is better able to recover at least 366/527 attributes compared to concurrent methods.}
    \label{fig:audioset_heatmap}
\end{figure}

\autoref{tab:probe-loss-all} reports the median probe loss for each task. In~\autoref{fig:audioset_heatmap}, we perform attribute-wise comparisons on AST-Audioset, the studied task comprising  the largest number of attributes. The numbers represent how many times the method of each row better recovers the attributes than the methods on the column. For instance, last row show that our method  attributes of Audioset.

Our method is able to better recover at least 366/527 attributes (69.4\%) than other methods. Performing a Wilcoxon signed-rank test, we obtain a statistic of 106584 with a p-value of $1.7\times 10^{-26}$, rejecting the null hypothesis thus proving the significance of those probe loss results.

In a similar fashion on CLIP-WikiArt, our method reaches a lower probe loss than TopKSAE on 140/167 attributes (83.8\%, even with TopKSAE reaching a lower probe loss on the ``style'' attributes), obtaining a test statistic of 12671 and a p-value of $7.9 \times 10^{-20}$, rejecting the null hypothesis. 

\section{Appendix: Statistical Significance of MPPC}\label{app:mppc}
The Maximum Pairwise Pearson Correlation (MPPC) was proposed by \citet{wang2025towards} as  a similarity indicator between models.
\subsection{Definition of MPPC}

To compare two sets of extracted concepts $A$ and $B$,
$\rho_i^{A\rightarrow B}$ is defined as the maximum pairwise Pearson
correlation between the $i$-th concept of $A$ and all concepts of $B$.
With $\bm{f}_i^A$ the vector containing values for each sample for the
$i$-th concepts of $A$, $\mu_i^A$ and $\sigma_i^A$ its mean and standard
deviation (respectively for $\bm{f}_j^B$, $\mu_j^B$ and $\sigma_j^B$):

\begin{equation}
    \rho_i^{A\rightarrow B} = \max_{j} \frac{\mathbb{E}[(\bm{f}_i^A - \mu_i^A)(\bm{f}_j^B - \mu_j^B)]}{\sigma_i^A \sigma_j^B}
    \label{rho-eq}
\end{equation}

Then, $\textit{MPPC}^{A \rightarrow B}$ is defined as the arithmetic
mean of $\rho_i^{A\rightarrow B}$ over all $i$, quantifying the extent
to which the concepts in $A$ are represented in $B$. In order to measure
consistency of a concept extraction method, we measure \textit{MPPC} 10
times between sets of concepts extracted with different random seeds, and
report the average. Therefore, a MPPC closer to 1 indicates a higher consistency.

\subsection{Statistical significance in our case}
With $\rho_i$ the maximum pairwise coefficient (Eq. ~\ref{rho-eq}) for $k$ target features of length $N$, and $H_0$ the hypothesis of features having no linear relationship. Using the Fischer z-transformation \cite{fisher1915frequency}

$$
z = artanh(r) \sim \mathcal{k}(0, \frac{1}{\sqrt{N - 3}})
$$
$$
\mathbb{P}(\max_{i}(r_i) > x) = 1 - \mathbb{P}( r \le x)^k
$$
$$
\mathbb{P}(\rho_i > x) = \mathbb{P}(\max_{i}(z_i) > artanh(x)) 
$$
$$
\mathbb{P}(\rho_i > x) = 1 - \Phi(artanh(x) \sqrt{N-3}) ^k
$$

With $k = 6144$ (corresponding to the main experiments), and $L=10000$ being largely lower than the size of the most used datasets, we obtain $\mathbb{P}(\rho_i > 0.3) \approx 10^{-206}$ , thus reject $H_0$.

\section{Appendix: Qualitative Examples of Extracted Concepts}
\label{app:concepts}

\begin{figure}[htbp]
    \centering
    \begin{subfigure}[b]{0.32\textwidth}
        \fbox{\includegraphics[width=\textwidth]{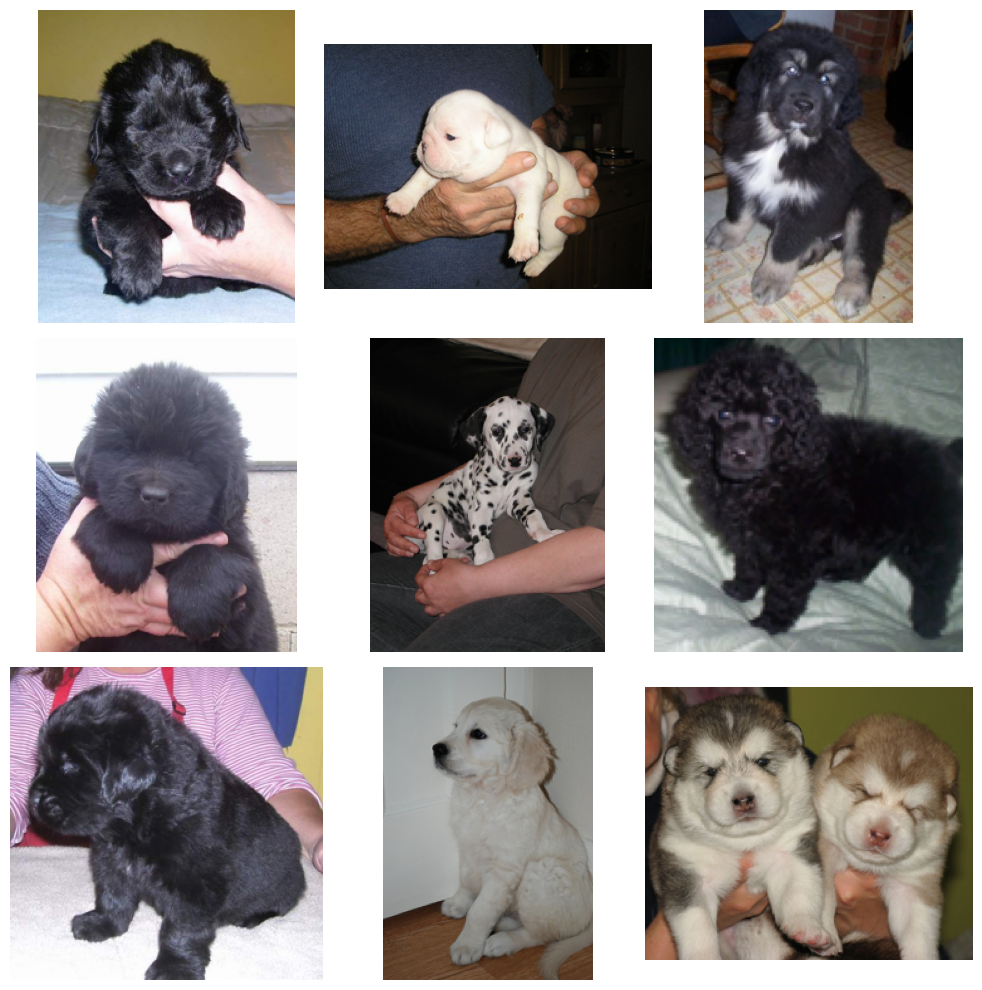}}
        \caption{Puppies}
        \label{fig:puppies}
    \end{subfigure}
    \hfill
    \begin{subfigure}[b]{0.32\textwidth}
        \fbox{\includegraphics[width=\textwidth]{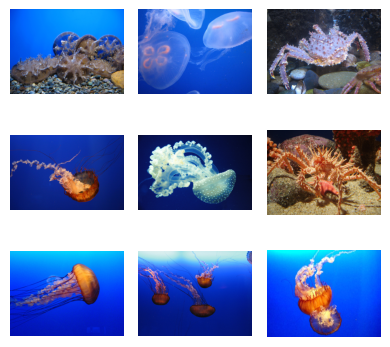}}
        \caption{Underwater Animals}
        \label{fig:underwater}
    \end{subfigure}
    \hfill
    \begin{subfigure}[b]{0.32\textwidth}
        \fbox{\includegraphics[width=\textwidth]{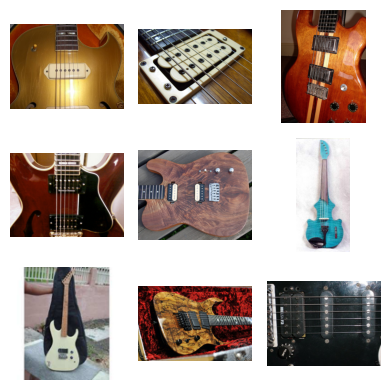}}
        \caption{Guitars}
        \label{fig:guitars}
    \end{subfigure}
    \begin{subfigure}[b]{0.32\textwidth}
        \fbox{\includegraphics[width=\textwidth]{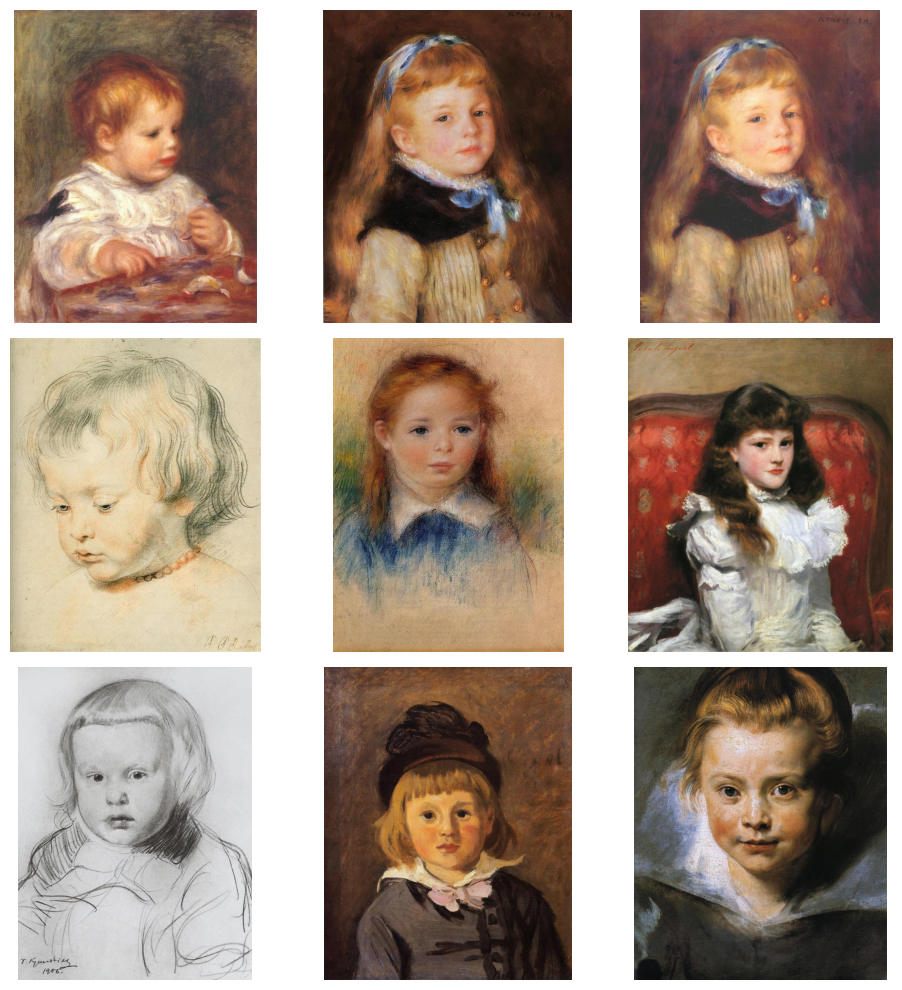}}
        \caption{Portraits of children}
        \label{fig:children}
    \end{subfigure}
    \hfill
    \begin{subfigure}[b]{0.32\textwidth}
        \fbox{\includegraphics[width=\textwidth]{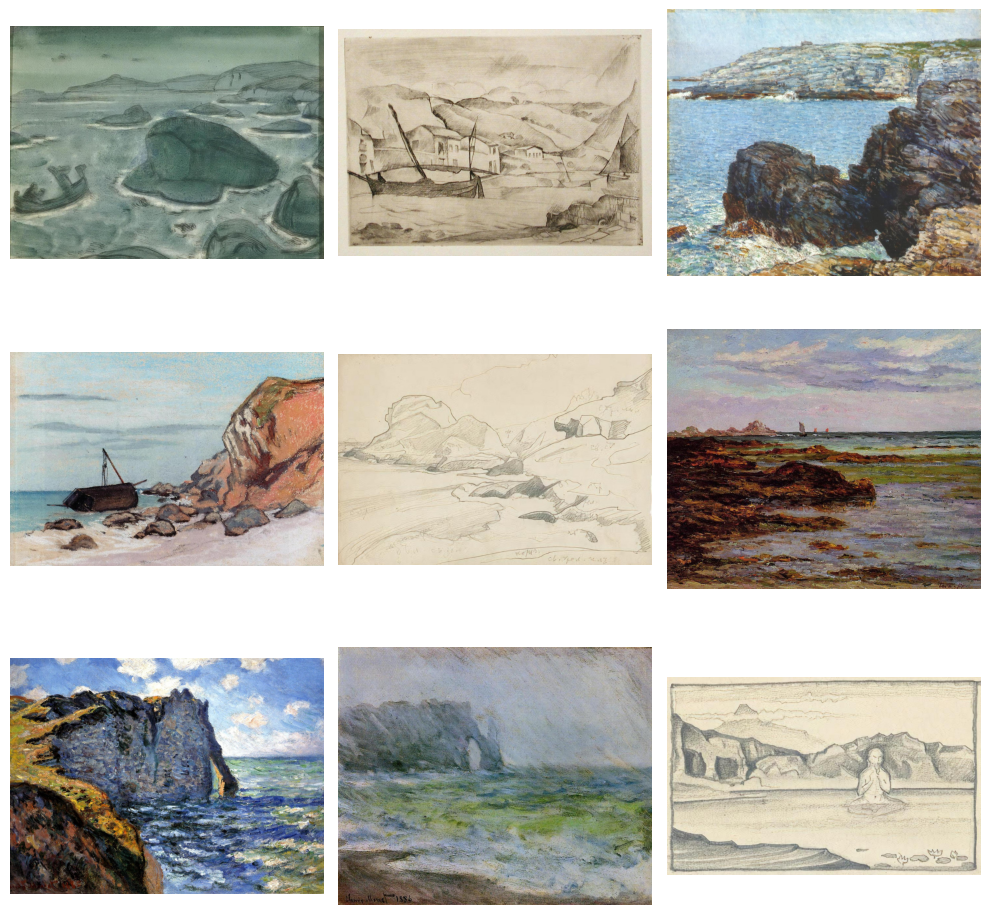}}
        \caption{Paintings of cliffs}
        \label{fig:cliffs}
    \end{subfigure}
    \hfill
    \begin{subfigure}[b]{0.32\textwidth}
        \fbox{\includegraphics[width=\textwidth]{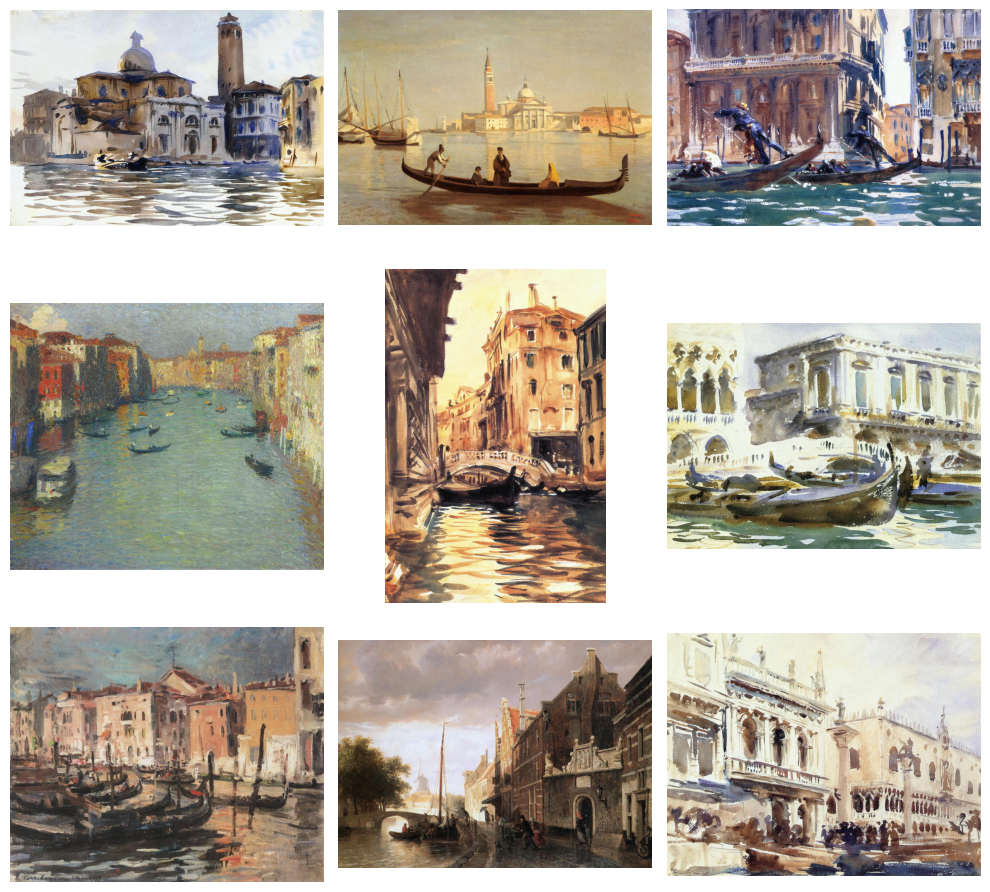}}
        \caption{Paintings of Venice}
        \label{fig:venice}
    \end{subfigure}
    \begin{subfigure}[b]{0.32\textwidth}
        \fbox{\includegraphics[width=\textwidth]{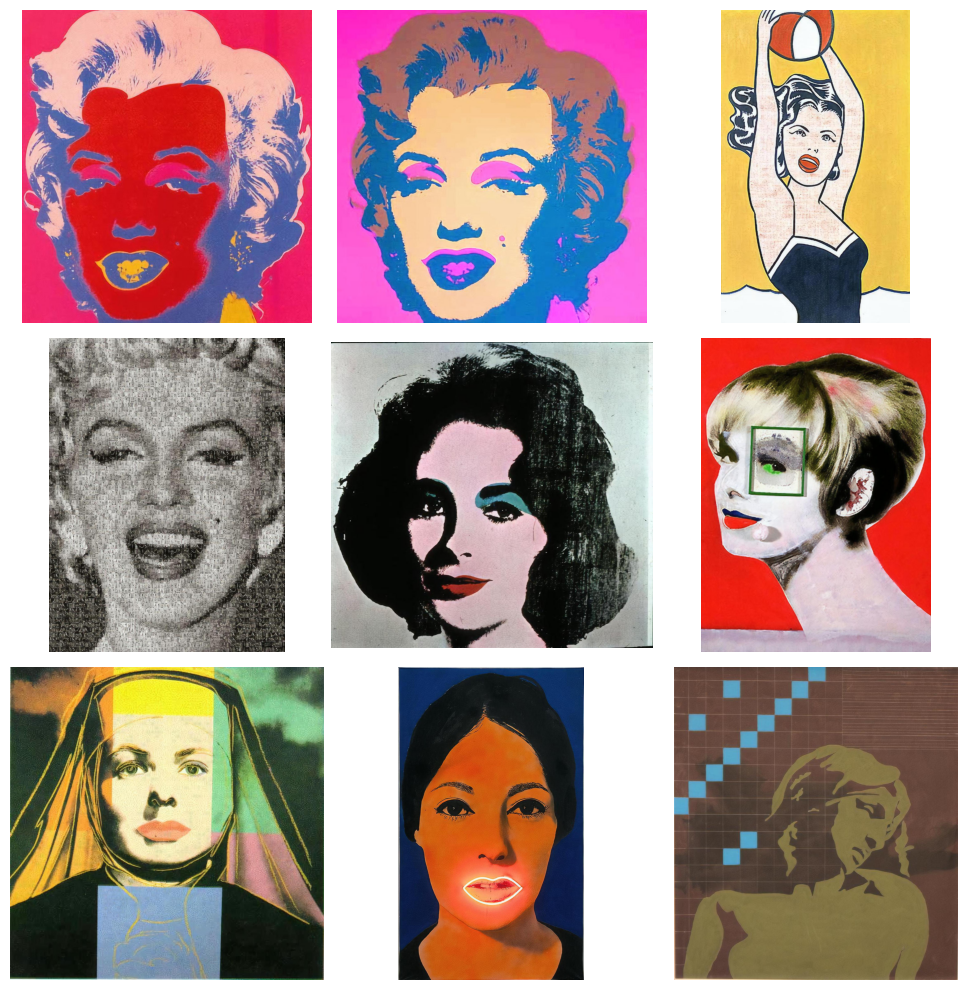}}
        \caption{Popart}
        \label{fig:popart}
    \end{subfigure}
    \hfill
    \begin{subfigure}[b]{0.32\textwidth}
        \fbox{\includegraphics[width=\textwidth]{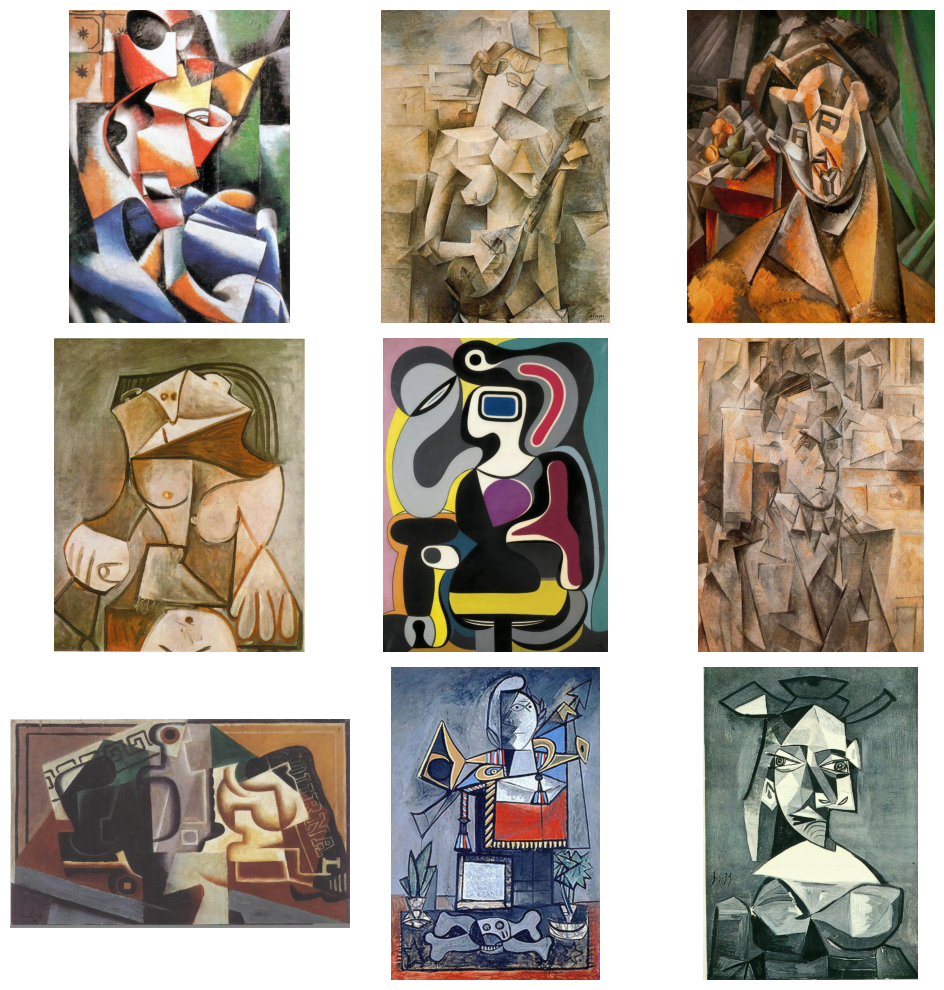}}
        \caption{Cubist paintings}
        \label{fig:cubism}
    \end{subfigure}
    \hfill
    \begin{subfigure}[b]{0.32\textwidth}
        \fbox{\includegraphics[width=\textwidth]{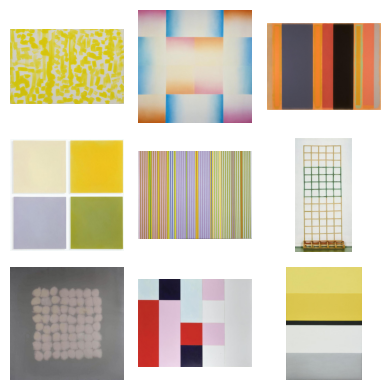}}
        \caption{Minimalist paintings}
        \label{fig:minimalism}
    \end{subfigure}
    \caption{Nine examples of visual concepts extracted from CLIP, over ImageNet and WikiArt. Representing the 9 images with the
    highest activations for each.}
    \label{fig:examples_img}
\end{figure}

\paragraph{Image Concepts} We present in \figref{fig:examples_img} three examples of concepts extracted from image models, from different datasets. The concepts are represented by the images with their nine highest activations. The name of the concepts are empirically set from the images. Displayed concepts are extracted from CLIP's activations, with ~\cref{fig:puppies,fig:underwater,fig:guitars} extracted from ImNet, ~\cref{fig:children,fig:cliffs,fig:venice} from WikiArt and corresponding to paintings content, and ~\cref{fig:popart,fig:cubism,fig:minimalism} corresponding to artistic styles.

\paragraph{Text Concepts}

In \autoref{tab:examples_txt} and~\autoref{tab:examples_txt_2}, we represent 3 textual concepts. For each
concept, we display the 3 sentences containing the highest token-wise
concept values, and underline tokens among the top-100. 

\begin{table}[htbp]
    \centering
    \caption{Examples of textual concepts extracted from DeBERTa on CoNLL-2003. {\color{black} Each column is a concept with three representative texts. Concept names are ours.}}
    \label{tab:examples_txt}
    \renewcommand{\arraystretch}{1.3} 
    \begin{tabular}{p{0.3\textwidth} p{0.3\textwidth} p{0.3\textwidth}}
        \toprule
        \textbf{Sports Achievements} & \textbf{Last Names} & \textbf{Nationalities} \\
        \midrule
        \textit{Seven athletes went into Friday's penultimate meeting of the series with a chance of winning the \underline{prize}.} & 
        \textit{Katarina \underline{Studen}ikova (Slovakia) beat 6- Karina H\underline{abs}udova.} &
        \textit{One \underline{Romanian} passenger was killed, and 14 others were injured on Thursday when a \underline{Romanian}-registered bus collided with a \underline{Bulgarian} one in northern Bulgaria, police said.} \\
        \midrule
        \textit{Russia's double Olympic champion Svetlana Masterkova smashed her second world \underline{record} in just 10 days on Friday when she bettered the \underline{mark} for the women's 1,000 metres.} &
        \textit{Hendrik D\underline{reek}man (Germany) vs. Greg R\underline{used}ski (Britain).} &
        \textit{He said a \underline{Turkish} civil aviation authority official had made the same point and he noted that a \underline{Turkish} plane had a similar accident there in 1994.} \\
        \midrule
        \textit{Jamaican veteran Merlene Ottey, who beat Devers in Zurich after just missing out on the \underline{gold medal} in Atlanta after a photo finish, had to settle for third place in 11.04.} &
        \textit{The Greek socialist party's executive bureau gave the green light to Prime Minister Costas Simitis to call snap elections, its general secretary Costas Sk\underline{andal}idis told reporters.} &
        \textit{A \underline{Polish} school girl blackmailed two women with anonymous letters threatening death and later explained that she needed money for textbooks, police said on Thursday.} \\
        \bottomrule
    \end{tabular}
\end{table}

\begin{table}[htbp]
    \centering
    \caption{Additional Examples of textual concepts extracted from DeBERTa on CoNLL-2003. {\color{black} Each column is a concept with three representative texts. Concept names are ours.}}
    \label{tab:examples_txt_2}
    \renewcommand{\arraystretch}{1.3} 
    \begin{tabular}{p{0.3\textwidth} p{0.3\textwidth} p{0.3\textwidth}}
        \toprule
        \textbf{Years from the 1990's} & \textbf{Age} & \textbf{Geopolitical Evolutions} \\
        \midrule
        \textit{West lake, arrested in December \underline{1993} and charged with heroin trafficking , sawed the iron grill off his cell window}
        & 
        \textit{Machado, \underline{19}, flew to Los Angeles after slipping away from the New Mexico desert town of Las Cruces}
        &
        \textit{Peruvian guerrillas killed one man and \underline{took} eight people hostage after \textbf{taking over} a village in the country' s northeastern jungle}  \\
        \midrule
        \textit{Since taking over as captain from Ne ale Fraser in \underline{1994}, Newcombe' s record in tandem with Roche, his former doubles partner, has been three wins and three losses.}
        &
        \textit{The \underline{13} - \underline{year} - \underline{old} girl tried to extract 60 and 70 zlotys ( \$22 and \$26 ) from two residents of Sierakowice by threatening to take their lives.}
        &
        \textit{[...] is ready at any time without preconditions to \underline{enter} peace negotiations} \\
        \midrule
        \textit{The bullish comments for the coming year soothed analysts and most shareholders , who were disappointed by the lower than expected profit for \underline{1995}/96.}
        &
        \textit{On Tuesday night , Kevorkian attended the death of Louise Siebens, a \underline{76}-\underline{year}-\underline{old} Texas woman with amyotrophic lateral sclerosis}
        &
        \textit{[..] that is to \underline{end} the state of hostility} \\
        \bottomrule
    \end{tabular}
\end{table}



\section{Appendix: Additional Steering Examples}

\paragraph{Textual Concept: Baseball} Extracted from DeBERTa, over CoNLL-2003. Enhancing this concept (positive values of alpha) causes replacement of any sport-specific terms (football, basketball) by their baseball equivalent. Those changes affect mentions of teams, leagues and scoring methods.
\begin{itemize}
    \item ($\pm$ 0) The best sport is basketball, NBA is the best $\rightarrow$
    (+3.75) The best sport is baseball, MLB is the best
    \item ($\pm$ 0) He scored 3 touchdowns in the first half $\rightarrow$ (+4.5) He scored 3 RBI in the first inning
    \item ($\pm$ 0) The New York Knicks beat the Los Angeles Lakers $\rightarrow$ (+3.75) The New York Yankees beat the Los Angeles Dodgers
\end{itemize}

\begin{figure}
    \centering
    \includegraphics[width=\textwidth]{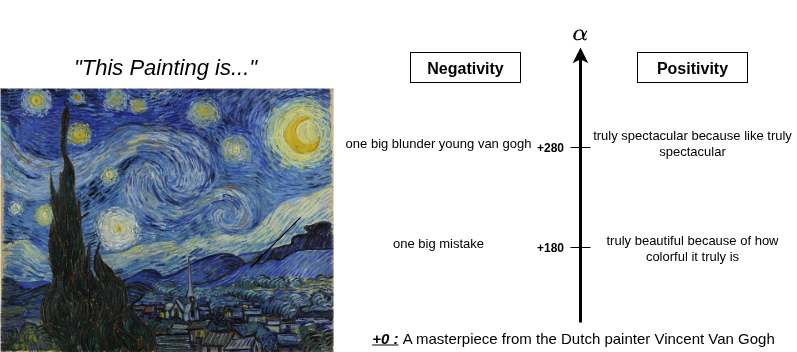}
    \caption{Steering Gemma3 captioning of \textit{The Starry Night}, by Vincent Van Gogh, upon 2 concepts corresponding to positivity and negativity.}
    \label{fig:steergemma}
\end{figure}

\paragraph{Steering Gemma3 Image Captioning}
Our method can extract concepts from large decoder models. From the text decoder of  Gemma3-4B-PT~\cite{team2025gemma}, we extract concepts over the IMDB dataset. We steer two concepts identified as corresponding to positivity/negativity during image captioning, see~\autoref{fig:steergemma}.

{\color{black}
\section{Quadratic Extension of Deleuzian Concepts}\label{app:quadratic_extension}

As our Deleuzian approach is analog to a Linear Discriminant Analysis (LDA) as per \autoref{sec:lda}, it makes hypothesis about isotropic distribution of concepts in a model's activation. However, we can derive an  extension of the  Deleuzian method, that does not make those hypothesis, analog with \textit{Quadratic Discriminant Analysis}. The aim is to extract discriminant functions $\delta_i : \mathbb{R^d} \rightarrow \mathbb{R}$ from neural networks' activations. Each function $\delta_i$ must then correspond to an interpretable concept. 

\subsection{Discriminant Function $\delta$}

A discriminant function $\delta$ is defined from a randomly sampled pair of samples $x_i, x_j$
the linear Deleuzian method considers linear concepts.
$$
\delta(x) = b^Tx
$$

With $b = x_i - x_j$. Such formulation is equivalent to a linear discriminant analysis (with hypothesis of homoscedasticity).

A Quadratic discriminant analysis (QDA) would require

$$
\delta(x) = -\frac{1}{2}x^T A x + b^Tx + c
$$

with $A = \Sigma_i^{-1} - \Sigma_j^{-1}$ 
and $b = \Sigma_i^{-1}\mu_i - \Sigma_j^{-1}\mu_j$
and $c = -\frac{1}{2}(\mu_i^T\Sigma_i^{-1}\mu_i - \mu_j^T\Sigma_j^{-1}\mu_j) - \frac{1}{2}\log \frac{|\Sigma_i|}{|\Sigma_j|}$ 
$c$ is constant, and only affecting thresholding, but not the geometry of $\delta$. Therefore we consider it neglectible.

To form covariance matrices $\Sigma_i, \Sigma_j$, we use Ledoit-Wolf shrinkage on the 50-neighborhoods of $x_i$ and $x_j$. Shrinkage methods are necessary to approximate covariance matrices on small datasets or with large dimensions. With $S$ the sample covariance, and $F = \frac{Tr(S)}{d}$ it estimates the optimal $\alpha_{LW}$

$$
\alpha_{LW} = \frac{(\frac{1}{T}\sum_{t=1}^{T} ||(x_t-\bar{x})(x_t-\bar{x})^T - S||^2_F)-(tr((S-F)\cdot \frac{1}{T}\sum_{t=1}^{T} (x_t-\bar{x})(x_t-\bar{x})^T - S}{tr((S-F)²)}
$$

$$
\hat{\Sigma} = (1 - \alpha) S  + \alpha_{LW} F
$$
We then use $\hat{\Sigma}_i, \hat{\Sigma}_j$ as covariance matrix for neighborhoods of $x_i$ and $x_j$ .

\subsection{Concept selection}

After extraction of $N$ candidate discriminant $\delta_i$, the linear Deleuzian method is restrained to $k$ concepts by performing feature weighted KMeans clustering.

\paragraph{Distance}
As Deleuzian concepts $\delta_i$ are linear, and only defined by a discriminant vector $b \in \mathbb{R}^d$, a trivial distance between two discriminant functions $\delta_i, \delta_j$ is the Euclidean distance between those vectors $||b_i, b_j||_2$. However, such metric cannot be computed on quadratic concepts having a more complex formulation.
From two discriminants $\delta_i, \delta_j$, we define the functional $L_w^{2}$ metric as 

$$
D_w^{2}(\delta_i, \delta_j) = \int_{\mathbb{R}^d}(\delta_i(x) - \delta_j(x))^{2}w(x)dx
$$
or in probalistic terms

$$
D_w^{2}(\delta_i, \delta_j) = \mathbb{E}_{x \sim w}[(\delta_i(x) - \delta_j(x))²]
$$
Using the support measure $w(x) = \mathcal{N}(0, I)$. $w(x)$ represents prior belief that data should follow a zero-mean, isotropic gaussian distribution. Using $w'(x) = \mathcal{N}(0, \alpha I), \alpha \in \mathbb{R}^{+}$ would only cause uniform scaling of $D^{2}$, without modifying the underlying geometry.
Noting $\Delta A = A_i - A_j$ and $\Delta b = b_i - b_j$, we have
$$
D_w^{2}(\delta_i, \delta_j) = \mathbb{E}_{x \sim w}[(-\frac{1}{2}x^T\Delta A x + \Delta b ^T x)^{2}]
$$
Developping, we consider the odd moments to vanish (as $w$ is a zero-mean gaussian). Therefore we obtain

$$
D_w^{2}(\delta_i, \delta_j) = \frac{1}{4}\mathbb{E}[(x^T\Delta A x)^{2}] + \mathbb{E}[(\Delta b ^T x)^{2}]
$$
Simplifying the linear term, we obtain

$$
\mathbb{E}[(\Delta b ^T x)^{2} = \Delta b ^T \mathbb{E}[xx^T] \Delta b
$$
As $x \sim \mathcal{N}(0, I)$, $\mathbb{E}[xx^T] = I_d$. Thus

$$
\mathbb{E}[(\Delta b ^T x)^{2}] = \Delta b ^T I_d \Delta b = ||\Delta b||^{2}
$$ Concerning the quadratic term, because $x \sim \mathcal{N}(0, I)$ we have
$$
\mathbb{E}[(x^T\Delta A x)^{2}] = 2Tr(\Delta A ^{2}) + Tr(\Delta A)^{2}
$$Quadratic parameters $A_i$ and $A_j$ are differences of covariance matrices formed with Ledoit-Wolf shrinking. Therefore, their diagonals are most likely similar, and dominated by constant isotropic offset. Then, we consider  $Tr(\Delta A)^{2} = Tr(A_i-A_j)^{2} \approx 0$, and we get

$$
\mathbb{E}[(x^T\Delta A x)^{2}] = 2Tr(\Delta A ^{2}) = 2||\Delta A||_F^{2}
$$

Therefore, our functional $L_w^{2}$ distance stands as follows :

$$
D^{2}(\delta_i,\delta_j) = \frac{1}{2} ||A_i-A_j||_F^{2} + ||b_i-b_j||^{2}
$$

\paragraph{Centroids Recomputation}
Once we have defined a functional distance, the main crucial step of KMeans clustering is the iteratice centroids recomputation. Each $\delta_i$ is assigned to its closest centroid $\bar{C}$, then $\bar{C}$ is recomputed in order to minimize within cluster distortion.
We recompute $\bar{C}$ (with parameters $\bar{A}, \bar{b}$) using the Fréchet mean upon our functional $L_w^{2}$ distance

$$
\bar{C} = \text{argmin}_\delta \sum_iw_i D²(\delta,\delta_i)
$$

$$
\bar{C} = \text{argmin}_\delta \sum_i w_i (\frac{1}{2}||A - A_i||_F^{2} + ||b - b_i||^{2})
$$

with ponderation weights $w_i$ (usually uniform, for unweighted mean)
Using the $A$ and $b$ derivatives of $\bar{C}$ to minimize distortion :

$$
\frac{\partial}{\partial A} \sum_i (\frac{1}{2}||A-A_i||_F^{2})=0 \implies \sum_i w_i(A - A_i) = 0 \implies A = \sum_iw_iA_i
$$
$$
\frac{\partial}{\partial b} \sum_i (\frac{1}{2}||b-b_i||^{2})=0 \implies \sum_i w_i(b - b_i) = 0 \implies b = \sum_iw_ib_i
$$

Therefore, we use $\bar{A} = \sum_i w_iA_i$ and $\bar{b} = \sum_i w_ib_i$ as parameters of the centroid $\bar{C}$

\subsection{Results and discussion about quadratic extension}
The obtained method is an exact generalization of our linear Deleuzian method to quadratic functions. \autoref{tab:qda} demonstrates that this extension reaches probe loss results better than SAE-based methods on CLIP-WikiArt, but does not outperform Linear Deleuzian concepts that is presented in the main paper. Such results may be due to the need to estimate covariance matrices on very high dimensional data.

\begin{table}[tbh]
\centering
\caption{\color{black}Results of  the Quadratic extension of Deleuzian concepts on CLIP-WikiArt}
\label{tab:qda}
\begin{tabular}{llll}
\hline
\multicolumn{1}{c}{\multirow{3}{*}{Methods}} & \multicolumn{3}{c}{CLIP}                            \\ \cline{2-4} 
\multicolumn{1}{c}{}                         & \multicolumn{3}{c}{WikiArt}                         \\ \cline{2-4} 
\multicolumn{1}{c}{}                         & Artist          & Style           & Genre           \\ \hline
Van-SAE                                      & 0.0137          & \textbf{0.0558} & 0.1531          \\
Tk-SAE                                       & 0.0125          & \textit{0.0558} & 0.1360          \\
Linear-Deleuzian (Main Method)               & \textbf{0.0119} & \textit{0.0560} & \textbf{0.1230} \\
\color{black} Quadratic-Deleuzian (Extension)              & \color{black}\textit{0.0124} & \color{black}0.6160          & \color{black}\textit{0.1305} \\ \hline
\end{tabular}
\end{table}

}
\section{Appendix: LLM usage}
Beyond the usage of LLM described in the paper, that is part of the study, we used commercial services to polish the writting: find synonyms, rephrase sentences.

%% file: iclr2026_conference.bib
@inproceedings{gao2024scaling,
  title={Scaling and evaluating sparse autoencoders},
  author={Gao, Leo and la Tour, Tom Dupr{\'e} and Tillman, Henk and Goh, Gabriel and Troll, Rajan and Radford, Alec and Sutskever, Ilya and Leike, Jan and Wu, Jeffrey},
  booktitle={International Conference on Representation Learning (ICLR)},
  year={2025}
}

@inproceedings{bussmann2024batchtopk,
  title={BatchTopK Sparse Autoencoders},
  author={Bussmann, Bart and Leask, Patrick and Nanda, Neel},
  booktitle={NeurIPS 2024 Workshop on Scientific Methods for Understanding Deep Learning},
  year={2024}
}

@article{rajamanoharan2024jumping,
  title={Jumping ahead: Improving reconstruction fidelity with jumprelu sparse autoencoders},
  author={Rajamanoharan, Senthooran and Lieberum, Tom and Sonnerat, Nicolas and Conmy, Arthur and Varma, Vikrant and Kram{\'a}r, J{\'a}nos and Nanda, Neel},
  journal={arXiv preprint arXiv:2407.14435},
  year={2024}
}

@inproceedings{bussmannlearning,
  title={Learning Multi-Level Features with Matryoshka Sparse Autoencoders},
  author={Bussmann, Bart and Nabeshima, Noa and Karvonen, Adam and Nanda, Neel},
  booktitle={Forty-second International Conference on Machine Learning},
  year={2025}
}

@inproceedings{zaigrajewinterpreting,
  title={Interpreting CLIP with Hierarchical Sparse Autoencoders},
  author={Zaigrajew, Vladimir and Baniecki, Hubert and Biecek, Przemyslaw},
  booktitle={Forty-second International Conference on Machine Learning},
  year={2025}
}

@article{fisher36lda,
  author = {Fisher, R. A.},
  journal = {Annals of Eugenics},
  number = 7,
  pages = {179-188},
  title = {The Use of Multiple Measurements in Taxonomic Problems},
  volume = 7,
  year = 1936
}

@book{mclachlan2005discriminant,
  title={Discriminant analysis and statistical pattern recognition},
  author={McLachlan, Geoffrey J},
  year={2005},
  publisher={John Wiley \& Sons}
}

@article{rao1948utilization,
  title={The utilization of multiple measurements in problems of biological classification},
  author={Rao, C Radhakrishna},
  journal={Journal of the Royal Statistical Society. Series B (Methodological)},
  volume={10},
  number={2},
  pages={159--203},
  year={1948},
  publisher={JSTOR}
}

@article{ahdesmaki2010feature,
  title={Feature selection in omics prediction problems using cat scores and false nondiscovery rate control},
  author={Ahdesm{\"a}ki, Miika and Strimmer, Korbinian},
  journal = {Annals of Applied Statistics},
  volume = {4},
  number = {1},
  pages = {503-519},
  year={2010}
}

@article{bricken2023towards,
  title={Towards monosemanticity: Decomposing language models with dictionary learning},
  author={Bricken, Trenton and Templeton, Adly and Batson, Joshua and Chen, Brian and Jermyn, Adam and Conerly, Tom and Turner, Nick and Anil, Cem and Denison, Carson and Askell, Amanda and others},
  journal={Transformer Circuits Thread},
  volume={2},
  year={2023}
}

@article{cunningham2023sparse,
  title={Sparse autoencoders find highly interpretable features in language models},
  author={Cunningham, Hoagy and Ewart, Aidan and Riggs, Logan and Huben, Robert and Sharkey, Lee},
  journal={arXiv preprint arXiv:2309.08600},
  year={2023}
}

@article{zeng2019cs,
  title={CS Sparse K-means: An Algorithm for Cluster-Specific Feature Selection in High-Dimensional Clustering},
  author={Zeng, Xiangrui and Zheng, Hongyu},
  journal={arXiv preprint arXiv:1909.12384},
  year={2019}
}

@article{lloyd1982least,
  title={Least squares quantization in PCM},
  author={Lloyd, Stuart},
  journal={IEEE transactions on information theory},
  volume={28},
  number={2},
  pages={129--137},
  year={1982},
  publisher={IEEE}
}

@inproceedings{
wang2025towards,
title={Towards Universality: Studying Mechanistic Similarity Across Language Model Architectures},
author={Junxuan Wang and Xuyang Ge and Wentao Shu and Qiong Tang and Yunhua Zhou and Zhengfu He and Xipeng Qiu},
booktitle={The Thirteenth International Conference on Learning Representations},
year={2025},
url={https://openreview.net/forum?id=2J18i8T0oI}
}

@article{sharkey2025open,
  title={Open problems in mechanistic interpretability},
  author={Sharkey, Lee and Chughtai, Bilal and Batson, Joshua and Lindsey, Jack and Wu, Jeff and Bushnaq, Lucius and Goldowsky-Dill, Nicholas and Heimersheim, Stefan and Ortega, Alejandro and Bloom, Joseph and others},
  journal={arXiv preprint arXiv:2501.16496},
  year={2025}
}

@book{deleuze1968différence,
  title={Diff{\'e}rence et r{\'e}p{\'e}tition},
  author={Deleuze, G.},
  isbn={9782130585299},
  lccn={74399404},
  series={Biblioth{\`e}que de philosophie contemporaine : Histoire de la philosophie et philosophie g{\'e}n{\'e}rale},
  url={https://books.google.gm/books?id=8lEwAAAAYAAJ},
  year={1968},
  publisher={Presses Universitaires de France}
}

@book{hegel1816wissenschaft,
  title={Wissenschaft der Logik: Die objective Logik},
  author={Hegel, Georg Wilhelm Friedrich},
  volume={2},
  year={1816},
  publisher={Johann Leonhard Schrag}
}

@misc{plato375republic_vi,
  title     = "The Republic -- Book VI",
  author    = "Plato",
  year      = "c. 375 BCE",
  address   = "Athens"
}

@misc{sartre1946existentialisme,
  title={L'existentialisme est un humanisme},
  author={Sartre, Jean-Paul and Elka{\"\i}m-Sartre, Arlette},
  year={1946},
  publisher={Nagel Paris}
}

@book{nietzsche1889gotzen,
  title={G{\"o}tzen-D{\"a}mmerung oder Wie man mit dem Hammer philosophirt},
  author={Nietzsche, Friedrich},
  year={1889},
  publisher={CG Naumann}
}

@article{imagenet15russakovsky,
    Author = {Olga Russakovsky and Jia Deng and Hao Su and Jonathan Krause and Sanjeev Satheesh and Sean Ma and Zhiheng Huang and Andrej Karpathy and Aditya Khosla and Michael Bernstein and Alexander C. Berg and Li Fei-Fei},
    Title = { {ImageNet Large Scale Visual Recognition Challenge} },
    Year = {2015},
    journal   = {International Journal of Computer Vision (IJCV)},
    doi = {10.1007/s11263-015-0816-y},
    volume={115},
    number={3},
    pages={211-252}
}

@InProceedings{maas-EtAl:2011:ACL-HLT2011,
  author    = {Maas, Andrew L.  and  Daly, Raymond E.  and  Pham, Peter T.  and  Huang, Dan  and  Ng, Andrew Y.  and  Potts, Christopher},
  title     = {Learning Word Vectors for Sentiment Analysis},
  booktitle = {Proceedings of the 49th Annual Meeting of the Association for Computational Linguistics: Human Language Technologies},
  month     = {June},
  year      = {2011},
  address   = {Portland, Oregon, USA},
  publisher = {Association for Computational Linguistics},
  pages     = {142--150},
  url       = {http://www.aclweb.org/anthology/P11-1015}
}

@inproceedings{tjong-kim-sang-de-meulder-2003-introduction,
    title = "Introduction to the {C}o{NLL}-2003 Shared Task: Language-Independent Named Entity Recognition",
    author = "Tjong Kim Sang, Erik F.  and
      De Meulder, Fien",
    booktitle = "Proceedings of the Seventh Conference on Natural Language Learning at {HLT}-{NAACL} 2003",
    year = "2003",
    url = "https://www.aclweb.org/anthology/W03-0419",
    pages = "142--147",
}

@inproceedings{jort_audioset_2017,
    title	= {Audio Set: An ontology and human-labeled dataset for audio events},
    author	= {Jort F. Gemmeke and Daniel P. W. Ellis and Dylan Freedman and Aren Jansen and Wade Lawrence and R. Channing Moore and Manoj Plakal and Marvin Ritter},
    year	= {2017},
    booktitle	= {Proc. IEEE ICASSP 2017},
    address	= {New Orleans, LA}
}

@InProceedings{radford2021learning,
  title =        {Learning Transferable Visual Models From Natural Language Supervision},
  author =       {Radford, Alec and Kim, Jong Wook and Hallacy, Chris and Ramesh, Aditya and Goh, Gabriel and Agarwal, Sandhini and Sastry, Girish and Askell, Amanda and Mishkin, Pamela and Clark, Jack and Krueger, Gretchen and Sutskever, Ilya},
  booktitle =    {Proceedings of the 38th International Conference on Machine Learning},
  pages =        {8748--8763},
  year =         {2021},
  editor =       {Meila, Marina and Zhang, Tong},
  volume =       {139},
  series =       {Proceedings of Machine Learning Research},
  month =        {18--24 Jul},
  publisher =    {PMLR},
  url =          {https://proceedings.mlr.press/v139/radford21a.html}
}

@article{oquab2023dinov2,
  title={Dinov2: Learning robust visual features without supervision},
  author={Oquab, Maxime and Darcet, Timoth{\'e}e and Moutakanni, Th{\'e}o and Vo, Huy and Szafraniec, Marc and Khalidov, Vasil and Fernandez, Pierre and Haziza, Daniel and Massa, Francisco and El-Nouby, Alaaeldin and others},
  journal={arXiv preprint arXiv:2304.07193},
  year={2023}
}

@inproceedings{he2021deberta,
title={{DeBERTa}: Decoding-enhanced {BERT} with Disentangled Attention},
author={Pengcheng He and Xiaodong Liu and Jianfeng Gao and Weizhu Chen},
booktitle={International Conference on Learning Representations},
year={2021},
url={https://openreview.net/forum?id=XPZIaotutsD}
}

@inproceedings{lewis2020bart,
  title={BART: Denoising Sequence-to-Sequence Pre-training for Natural Language Generation, Translation, and Comprehension},
  author={Lewis, Mike and Liu, Yinhan and Goyal, Naman and Ghazvininejad, Marjan and Mohamed, Abdelrahman and Levy, Omer and Stoyanov, Veselin and Zettlemoyer, Luke},
  booktitle={Proceedings of the 58th Annual Meeting of the Association for Computational Linguistics},
  pages={7871--7880},
  year={2020}
}

@article{gong2021ast,
  title={AST: Audio Spectrogram Transformer},
  author={Gong, Yuan and Chung, Yu-An and Glass, James},
  journal={Interspeech 2021},
  year={2021},
  publisher={ISCA}
}

@inproceedings{devlin-etal-2019-bert,
    title = "{BERT}: Pre-training of Deep Bidirectional Transformers for Language Understanding",
    author = "Devlin, Jacob  and
      Chang, Ming-Wei  and
      Lee, Kenton  and
      Toutanova, Kristina",
    editor = "Burstein, Jill  and
      Doran, Christy  and
      Solorio, Thamar",
    booktitle = "Proceedings of the 2019 Conference of the North {A}merican Chapter of the Association for Computational Linguistics: Human Language Technologies, Volume 1 (Long and Short Papers)",
    month = jun,
    year = "2019",
    address = "Minneapolis, Minnesota",
    publisher = "Association for Computational Linguistics",
    url = "https://aclanthology.org/N19-1423/",
    doi = "10.18653/v1/N19-1423",
    pages = "4171--4186"
}

@inproceedings{kohn2015s,
  title={What’s in an Embedding? Analyzing Word Embeddings through Multilingual Evaluation},
  author={K{\"o}hn, Arne},
  booktitle={Proceedings of the 2015 Conference on Empirical Methods in Natural Language Processing},
  pages={2067--2073},
  year={2015}
}

@inproceedings{gupta2015distributional,
  title={Distributional vectors encode referential attributes},
  author={Gupta, Abhijeet and Boleda, Gemma and Baroni, Marco and Pad{\'o}, Sebastian},
  booktitle={Proceedings of the 2015 Conference on Empirical Methods in Natural Language Processing},
  pages={12--21},
  year={2015}
}

@inproceedings{
marks2024the,
title={The Geometry of Truth: Emergent Linear Structure in Large Language Model Representations of True/False Datasets},
author={Samuel Marks and Max Tegmark},
booktitle={First Conference on Language Modeling},
year={2024},
url={https://openreview.net/forum?id=aajyHYjjsk}
}

@misc{
alain2017understanding,
title={Understanding intermediate layers using linear classifier probes},
author={Guillaume Alain and Yoshua Bengio},
year={2017},
url={https://openreview.net/forum?id=ryF7rTqgl}
}

@article{lovering2022evaluation,
  title={Evaluation beyond task performance: analyzing concepts in AlphaZero in Hex},
  author={Lovering, Charles and Forde, Jessica and Konidaris, George and Pavlick, Ellie and Littman, Michael},
  journal={Advances in neural information processing systems},
  volume={35},
  pages={25992--26006},
  year={2022}
}

@article{belinkov2022probing,
  title={Probing classifiers: Promises, shortcomings, and advances},
  author={Belinkov, Yonatan},
  journal={Computational Linguistics},
  volume={48},
  number={1},
  pages={207--219},
  year={2022},
  publisher={MIT Press One Broadway, 12th Floor, Cambridge, Massachusetts 02142, USA~…}
}

@inproceedings{burnsdiscovering,
  title={Discovering Latent Knowledge in Language Models Without Supervision},
  author={Burns, Collin and Ye, Haotian and Klein, Dan and Steinhardt, Jacob},
  booktitle={The Eleventh International Conference on Learning Representations},
  year={2023}
}

@article{lee2007sparse,
  title={Sparse deep belief net model for visual area V2},
  author={Lee, Honglak and Ekanadham, Chaitanya and Ng, Andrew},
  journal={Advances in neural information processing systems},
  volume={20},
  year={2007}
}

@article{dunefsky2024transcoders,
  title={Transcoders find interpretable llm feature circuits},
  author={Dunefsky, Jacob and Chlenski, Philippe and Nanda, Neel},
  journal={Advances in Neural Information Processing Systems},
  volume={37},
  pages={24375--24410},
  year={2024}
}

@article{Lindsey2024SparseCrosscoders,
  author = {Jack Lindsey and Adly Templeton and Jonathan Marcus and Thomas Conerly and Jared Batson and Chris Olah},
  title = {Sparse Crosscoders for Cross-Layer Features and Model Diffing},
  year = {2024},
  journal = {Transformer Circuits},
  url = {https://transformer-circuits.pub/2024/crosscoders/index.html},
}

@article{conmy2023towards,
  title={Towards automated circuit discovery for mechanistic interpretability},
  author={Conmy, Arthur and Mavor-Parker, Augustine and Lynch, Aengus and Heimersheim, Stefan and Garriga-Alonso, Adri{\`a}},
  journal={Advances in Neural Information Processing Systems},
  volume={36},
  pages={16318--16352},
  year={2023}
}

@inproceedings{
zhou2025llm,
title={{LLM} Neurosurgeon: Targeted Knowledge Removal in {LLM}s using Sparse Autoencoders},
author={Dylan Zhou and Kunal Patil and Yifan Sun and Karthik lakshmanan and Senthooran Rajamanoharan and Arthur Conmy},
booktitle={ICLR 2025 Workshop on Building Trust in Language Models and Applications},
year={2025},
url={https://openreview.net/forum?id=aeQeXlG2Pw}
}

@article{thasarathan2025universal,
  title={Universal Sparse Autoencoders: Interpretable Cross-Model Concept Alignment},
  author={Thasarathan, Harrish and Forsyth, Julian and Fel, Thomas and Kowal, Matthew and Derpanis, Konstantinos},
  journal={arXiv preprint arXiv:2502.03714},
  year={2025}
}

@article{fisher1915frequency,
  title={Frequency distribution of the values of the correlation coefficient in samples from an indefinitely large population},
  author={Fisher, Ronald A},
  journal={Biometrika},
  volume={10},
  number={4},
  pages={507--521},
  year={1915},
  publisher={JSTOR}
}

@inproceedings{roy2007effective,
  title={The effective rank: A measure of effective dimensionality},
  author={Roy, Olivier and Vetterli, Martin},
  booktitle={2007 15th European signal processing conference},
  pages={606--610},
  year={2007},
  organization={IEEE}
}

@inproceedings{skeanlayer,
  title={Layer by Layer: Uncovering Hidden Representations in Language Models},
  author={Skean, Oscar and Arefin, Md Rifat and Zhao, Dan and Patel, Niket Nikul and Naghiyev, Jalal and LeCun, Yann and Shwartz-Ziv, Ravid},
  booktitle={Forty-second International Conference on Machine Learning},
  year={2025}
}

@inproceedings{koh2020concept,
  title={Concept bottleneck models},
  author={Koh, Pang Wei and Nguyen, Thao and Tang, Yew Siang and Mussmann, Stephen and Pierson, Emma and Kim, Been and Liang, Percy},
  booktitle={International conference on machine learning},
  pages={5338--5348},
  year={2020},
  organization={PMLR}
}

@article{rajamanoharan2024improving,
  title={Improving sparse decomposition of language model activations with gated sparse autoencoders},
  author={Rajamanoharan, Senthooran and Conmy, Arthur and Smith, Lewis and Lieberum, Tom and Varma, Vikrant and Kramar, Janos and Shah, Rohin and Nanda, Neel},
  journal={Advances in Neural Information Processing Systems},
  volume={37},
  pages={775--818},
  year={2024}
}

@article{huang2005automated,
  title={Automated variable weighting in k-means type clustering},
  author={Huang, Joshua Zhexue and Ng, Michael K and Rong, Hongqiang and Li, Zichen},
  journal={IEEE transactions on pattern analysis and machine intelligence},
  volume={27},
  number={5},
  pages={657--668},
  year={2005},
  publisher={IEEE}
}

@article{milligan1980examination,
  title={An examination of the effect of six types of error perturbation on fifteen clustering algorithms},
  author={Milligan, Glenn W},
  journal={psychometrika},
  volume={45},
  number={3},
  pages={325--342},
  year={1980},
  publisher={Springer-Verlag}
}

@article{comon1994independent,
  title={Independent component analysis, a new concept?},
  author={Comon, Pierre},
  journal={Signal processing},
  volume={36},
  number={3},
  pages={287--314},
  year={1994},
  publisher={Elsevier}
}

@inproceedings{rao2024discover,
  title={Discover-then-name: Task-agnostic concept bottlenecks via automated concept discovery},
  author={Rao, Sukrut and Mahajan, Sweta and B{\"o}hle, Moritz and Schiele, Bernt},
  booktitle={European Conference on Computer Vision},
  pages={444--461},
  year={2024},
  organization={Springer}
}

@inproceedings{kim2018interpretability,
  title={Interpretability beyond feature attribution: Quantitative testing with concept activation vectors (tcav)},
  author={Kim, Been and Wattenberg, Martin and Gilmer, Justin and Cai, Carrie and Wexler, James and Viegas, Fernanda and others},
  booktitle={International conference on machine learning},
  pages={2668--2677},
  year={2018},
  organization={PMLR}
}

@article{ghorbani2019towards,
  title={Towards automatic concept-based explanations},
  author={Ghorbani, Amirata and Wexler, James and Zou, James Y and Kim, Been},
  journal={Advances in neural information processing systems},
  volume={32},
  year={2019}
}

@article{fel2023holistic,
  title={A holistic approach to unifying automatic concept extraction and concept importance estimation},
  author={Fel, Thomas and Boutin, Victor and B{\'e}thune, Louis and Cad{\`e}ne, R{\'e}mi and Moayeri, Mazda and And{\'e}ol, L{\'e}o and Chalvidal, Mathieu and Serre, Thomas},
  journal={Advances in Neural Information Processing Systems},
  volume={36},
  pages={54805--54818},
  year={2023}
}

@article{thomee2016yfcc,
author = {Thomee, Bart and Shamma, David A. and Friedland, Gerald and Elizalde, Benjamin and Ni, Karl and Poland, Douglas and Borth, Damian and Li, Li-Jia},
title = {YFCC100M: the new data in multimedia research},
year = {2016},
issue_date = {February 2016},
publisher = {Association for Computing Machinery},
address = {New York, NY, USA},
volume = {59},
number = {2},
issn = {0001-0782},
url = {https://doi.org/10.1145/2812802},
doi = {10.1145/2812802},
journal = {Commun. ACM},
month = jan,
pages = {64–73},
numpages = {10}
}

@misc{ilharco_gabriel_2021_5143773,
  author       = {Ilharco, Gabriel and
                  Wortsman, Mitchell and
                  Wightman, Ross and
                  Gordon, Cade and
                  Carlini, Nicholas and
                  Taori, Rohan and
                  Dave, Achal and
                  Shankar, Vaishaal and
                  Namkoong, Hongseok and
                  Miller, John and
                  Hajishirzi, Hannaneh and
                  Farhadi, Ali and
                  Schmidt, Ludwig},
  title        = {OpenCLIP},
  month        = jul,
  year         = 2021,
  publisher    = {Zenodo},
  version      = {0.1},
  doi          = {10.5281/zenodo.5143773},
  url          = {https://doi.org/10.5281/zenodo.5143773}
}

@misc{wikiart,
    author = {Peter Baylies},
    title = {WikiArt dataset},
    year = {2020},
    url = {https://www.kaggle.com/datasets/steubk/wikiart}
}

@article{team2025gemma,
  title={Gemma 3 technical report},
  author={Team, Gemma and Kamath, Aishwarya and Ferret, Johan and Pathak, Shreya and Vieillard, Nino and Merhej, Ramona and Perrin, Sarah and Matejovicova, Tatiana and Ram{\'e}, Alexandre and Rivi{\`e}re, Morgane and others},
  journal={arXiv preprint arXiv:2503.19786},
  year={2025}
}

@inproceedings{arik2020explaining,
  title={Explaining deep neural networks using unsupervised clustering},
  author={Arik, Sercan and Liu, Yu-Han},
  booktitle={Proc. Workshop Hum. Interpretability Mach. Learn},
  pages={377--389},
  year={2020}
}

@article{hyvarinen2000ica_algo_appli,
title = {Independent component analysis: algorithms and applications},
journal = {Neural Networks},
volume = {13},
number = {4},
pages = {411-430},
year = {2000},
issn = {0893-6080},
doi = {https://doi.org/10.1016/S0893-6080(00)00026-5},
url = {https://www.sciencedirect.com/science/article/pii/S0893608000000265},
author = {A. Hyvärinen and E. Oja}
}

@article{scikit-learn,
  title={Scikit-learn: Machine Learning in {P}ython},
  author={Pedregosa, F. and Varoquaux, G. and Gramfort, A. and Michel, V.
          and Thirion, B. and Grisel, O. and Blondel, M. and Prettenhofer, P.
          and Weiss, R. and Dubourg, V. and Vanderplas, J. and Passos, A. and
          Cournapeau, D. and Brucher, M. and Perrot, M. and Duchesnay, E.},
  journal={Journal of Machine Learning Research},
  volume={12},
  pages={2825--2830},
  year={2011}
}

@inproceedings{felarchetypal,
  title={Archetypal SAE: Adaptive and Stable Dictionary Learning for Concept Extraction in Large Vision Models},
  author={Fel, Thomas and Lubana, Ekdeep Singh and Prince, Jacob S and Kowal, Matthew and Boutin, Victor and Papadimitriou, Isabel and Wang, Binxu and Wattenberg, Martin and Ba, Demba E and Konkle, Talia},
  year = {2025},
  booktitle={Forty-second International Conference on Machine Learning (ICLR)}
}

@article{joseph2025prisma,
  title={Prisma: An Open Source Toolkit for Mechanistic Interpretability in Vision and Video},
  author={Joseph, Sonia and Suresh, Praneet and Hufe, Lorenz and Stevinson, Edward and Graham, Robert and Vadi, Yash and Bzdok, Danilo and Lapuschkin, Sebastian and Sharkey, Lee and Richards, Blake Aaron},
  journal={arXiv preprint arXiv:2504.19475},
  year={2025}
}

@article{wynen2018unsupervised,
  title={Unsupervised learning of artistic styles with archetypal style analysis},
  author={Wynen, Daan and Schmid, Cordelia and Mairal, Julien},
  journal={Advances in Neural Information Processing Systems},
  volume={31},
  year={2018}
}

@inproceedings{biderman2023pythia,
  title={Pythia: A suite for analyzing large language models across training and scaling},
  author={Biderman, Stella and Schoelkopf, Hailey and Anthony, Quentin Gregory and Bradley, Herbie and O’Brien, Kyle and Hallahan, Eric and Khan, Mohammad Aflah and Purohit, Shivanshu and Prashanth, USVSN Sai and Raff, Edward and others},
  booktitle={International Conference on Machine Learning},
  pages={2397--2430},
  year={2023},
  organization={PMLR}
}

@inproceedings{hindupur2025projecting,
title={Projecting Assumptions: The Duality Between Sparse Autoencoders and Concept Geometry},
author={Hindupur, Sai Sumedh R and Lubana, Ekdeep Singh and Fel, Thomas and Ba, Demba},
booktitle={ICML 2025 Workshop on Methods and Opportunities at Small Scale},
year={2025},
url={https://openreview.net/forum?id=AKaoBzhIIF}
}
